\documentclass[acmtog, authorversion=true]{acmart}

\usepackage{graphicx}
\usepackage{amsmath}
\usepackage{booktabs}
\usepackage{multirow}
\usepackage{dsfont}

\usepackage{overpic}
\usepackage{enumitem} 
\usepackage{overpic} 
\usepackage{color}
\usepackage{comment}

\definecolor{turquoise}{cmyk}{0.65,0,0.1,0.3}
\definecolor{purple}{rgb}{0.65,0,0.65}
\definecolor{dark_green}{rgb}{0, 0.5, 0}
\definecolor{orange}{rgb}{0.8, 0.6, 0.2}
\definecolor{red}{rgb}{0.9, 0.1, 0.1}
\definecolor{darkred}{rgb}{0.6, 0.1, 0.05}
\definecolor{blueish}{rgb}{0.0, 0.3, .6}
\definecolor{light_gray}{rgb}{0.7, 0.7, .7}
\definecolor{pink}{rgb}{1, 0, 1}
\definecolor{greyblue}{rgb}{0.25, 0.25, 1}

\newcommand{\rev}[1]{#1}

\DeclareMathOperator*{\argmin}{arg\,min}

\newcommand{\bd}{\mathbf{d}}
\newcommand{\be}{\mathbf{e}}
\newcommand{\bff}{\mathbf{f}}
\newcommand{\bg}{\mathbf{g}}

\newcommand{\bp}{\mathbf{p}}

\newcommand{\bx}{\mathbf{x}}

\newcommand{\bz}{\mathbf{z}}
\newcommand{\bA}{\mathbf{A}}
\newcommand{\bB}{\mathbf{B}}

\newcommand{\bF}{\mathbf{F}}

\newcommand{\bK}{\mathbf{K}}

\newcommand{\bP}{\mathbf{P}}
\newcommand{\bQ}{\mathbf{Q}}

\newcommand{\bV}{\mathbf{V}}

\newcommand{\bX}{\mathbf{X}}

\def\K{{\mathcal{K}}}

\def\N{{\mathcal{N}}}

\def\X{{\mathcal{X}}}

\def\Z{{\mathcal{Z}}}

\def\code#1{\texttt{#1}}
\usepackage{blindtext}

\newcommand{\vqarchitecture}{
\begin{table*}[tbp]
\centering
\caption{Architecture of the image encoder and decoder. Note that $H_\text{feat} = \frac{H_\text{im}}{16}$, $W_\text{feat} = \frac{W_\text{im}}{16}$.}
\begin{tabular}{c|c}
Encoder & Decoder \\
\toprule
$ \bX \in \mathbb{R}^{H_\text{im} \times W_\text{im} \times 3}$  & $\hat{\bF} \in \mathbb{R}^{H_\text{feat} \times W_\text{feat} \times 256}$ \\
Conv2D $\to \mathbb{R}^{H_\text{im} \times W_\text{im} \times 128} $ & Conv2D $\to \mathbb{R}^{H_\text{feat} \times W_\text{feat} \times 512} $ \\
$4 \times $ $\{$ Residual Block, Downsample Block$\}$ $\to \mathbb{R}^{H_\text{feat} \times W_\text{feat} \times 512} $ & Residual Block $\to \mathbb{R}^{H_\text{feat} \times W_\text{feat} \times 512} $ \\
Residual Block $\to \mathbb{R}^{H_\text{feat} \times W_\text{feat} \times 512} $ & Non-Local Block $\to \mathbb{R}^{H_\text{feat} \times W_\text{feat} \times 512} $ \\
Non-Local Block $\to \mathbb{R}^{H_\text{feat} \times W_\text{feat} \times 512} $ & Residual Block $\to \mathbb{R}^{H_\text{feat} \times W_\text{feat} \times 512} $ \\
Residual Block  $\to \mathbb{R}^{H_\text{feat} \times W_\text{feat} \times 512} $ & $4 \times $ $\{$ Residual Block, Upsample Block$\}$ $\to \mathbb{R}^{H_\text{im} \times W_\text{im} \times 128} $ \\
GroupNorm, Swish, Conv2D $\to \mathbb{R}^{H_\text{feat} \times W_\text{feat} \times 256}$  &  GroupNorm, Swish, Conv2D $\to \mathbb{R}^{H_\text{im} \times W_\text{im} \times 3}$\\
\bottomrule
\end{tabular}
\label{tab:vqarchitecture}
\end{table*}
}

\newcommand{\bartarchitecture}{
\begin{table*}[tbp]
\centering
\caption{Transformer hyperparameters. For every experiment, the number of total blocks $N$, the number of blocks in $\N(r)$, the number of blocks in $\K(r)$ are set to $64$, $3$, $3$ respectively. 
$n_E$ denotes the number of transformer layers in the bidirectional encoder, $n_D$ is the number of transformer layers in the autoregressive decoder, \# params is the number of transformer parameters, $n_h$ is the number of attention heads in the transformer, $\vert \Z \vert$ is the number of codebook entries, dropout is the dropout rate used for training the transformer, and $n_e$ is the token embedding dimensionality.
  }
\begin{tabular}{c   c c c  c c c c c c}
  Dataset & $n_{E}$ & $n_{D}$ & \# params $[M]$ & $n_h$ & $\vert \Z \vert$ & dropout & $n_e$ \\
\toprule
  Flickr-Landscape & 7 & 15 & 343 & 16 & 1024 & 0.0 & 1024   \\
  COCO-Stuff & 7 & 15 & 365 & 16 & 8192 & 0.0 & 1024   \\
  ADE20K & 7 & 10 & 269 & 16 & 4096 & 0.1 & 1024   \\
\bottomrule
\end{tabular}

\label{tab:bartarchitecture}
\end{table*}
}

\newcommand{\cocoade}{
\begin{table}[tbp]
\setlength{\tabcolsep}{2.5pt}
\centering
\caption{\rev{Quantitative evaluation on the COCO-Stuff and ADE20K datasets at $512 \times 512$ resolution.}}
\begin{tabular}{lccccccc}

\toprule
Dataset & Method & LPIPS $\downarrow$ & FID $\downarrow$ & SSIM $\uparrow$ & mIoU $\uparrow$ & accu $\uparrow$ & div $\uparrow$ \\

\midrule

COCO- & \emph{TT} & 0.237 & 20.2 & 0.820 & 36.9 & 51.7 & \textbf{0.192}  \\
Stuff & \emph{ASSET} & \textbf{0.194} & \textbf{14.7} & \textbf{0.845} & \textbf{43.4} & \textbf{58.7} & 0.156 \\
\midrule

\multirow{2}{*}{ADE20K} & \emph{TT } & 0.197 & 15.7 & 0.860 & 51.8 & 68.1 & \textbf{0.155}  \\
& \emph{ASSET} & \textbf{0.191} & \textbf{14.0} & \textbf{0.862} & \textbf{52.6} & \textbf{68.8} & 0.140 \\
\bottomrule
\end{tabular}

\label{tab:cocoade}
\end{table}
}

\AtBeginDocument{%
  \providecommand\BibTeX{{%
    \normalfont B\kern-0.5em{\scshape i\kern-0.25em b}\kern-0.8em\TeX}}}

\setcopyright{acmcopyright}
\copyrightyear{2022}
\acmDOI{10.1145/3528223.3530172}

\acmJournal{TOG}
\acmYear{2022}
\acmMonth{7}
\acmVolume{41}
\acmNumber{4}
\acmArticle{74}

\begin{document}

\title{ASSET: Autoregressive Semantic Scene Editing with Transformers at High Resolutions}

\author{Difan Liu}
\email{dliu@cs.umass.edu}
\affiliation{%
  \institution{UMass Amherst and Adobe Research}
  \country{USA}}
\orcid{0000-0001-5971-2748}

\author{Sandesh Shetty}
\email{sandeshshett@umass.edu}
\affiliation{%
  \institution{UMass Amherst}
  \country{USA}}
\orcid{0000-0003-0331-4809}

\author{Tobias Hinz}
\email{thinz@adobe.com}
\affiliation{%
  \institution{Adobe Research}
  \country{USA}}
\orcid{0000-0003-1354-1562}

\author{Matthew Fisher}
\email{techmatt@gmail.com}
\affiliation{%
  \institution{Adobe Research}
  \country{USA}}
\orcid{0000-0002-8908-3417}

\author{Richard Zhang}
\email{rizhang@adobe.com}
\affiliation{%
  \institution{Adobe Research}
  \country{USA}}

\author{Taesung Park}
\email{taesung_park@berkeley.edu}
\affiliation{%
  \institution{Adobe Research}
  \country{USA}}

\author{Evangelos Kalogerakis}
\email{kalo@cs.umass.edu}
\affiliation{%
  \institution{UMass Amherst}
  \country{USA}}
\orcid{0000-0002-5867-5735}
\renewcommand{\shortauthors}{Liu et al.}

\begin{abstract}
We present ASSET, a neural architecture for automatically modifying an input high-resolution image according to a user's edits on its semantic segmentation map. Our architecture is based on a transformer with a novel attention mechanism. Our key idea is to sparsify the transformer's attention matrix at high resolutions, guided by dense attention extracted at lower image resolutions.
While previous attention mechanisms are computationally too expensive for handling high-resolution images or are overly constrained within specific image regions hampering long-range interactions, our novel attention mechanism is both computationally efficient and effective. Our sparsified attention mechanism is able to capture long-range interactions and context, leading to synthesizing interesting phenomena in scenes, such as reflections of landscapes onto water or flora consistent with the rest of the landscape, that were not possible to generate reliably with previous convnets and transformer approaches. We present qualitative and quantitative results, along with user studies, demonstrating the effectiveness of our method. \rev{Our code and dataset are available at our project page: \mbox{\textcolor{blue}{\url{https://github.com/DifanLiu/ASSET}}}}
\end{abstract}

\ccsdesc[500]{Computing methodologies~Image processing}
\ccsdesc[300]{Computing methodologies~Neural networks}
\ccsdesc[100]{Computing methodologies~Image representations}

\keywords{image editing, neural networks, transformers, sparse attention}

\begin{teaserfigure}
\centering
\includegraphics[width=1.00\textwidth]{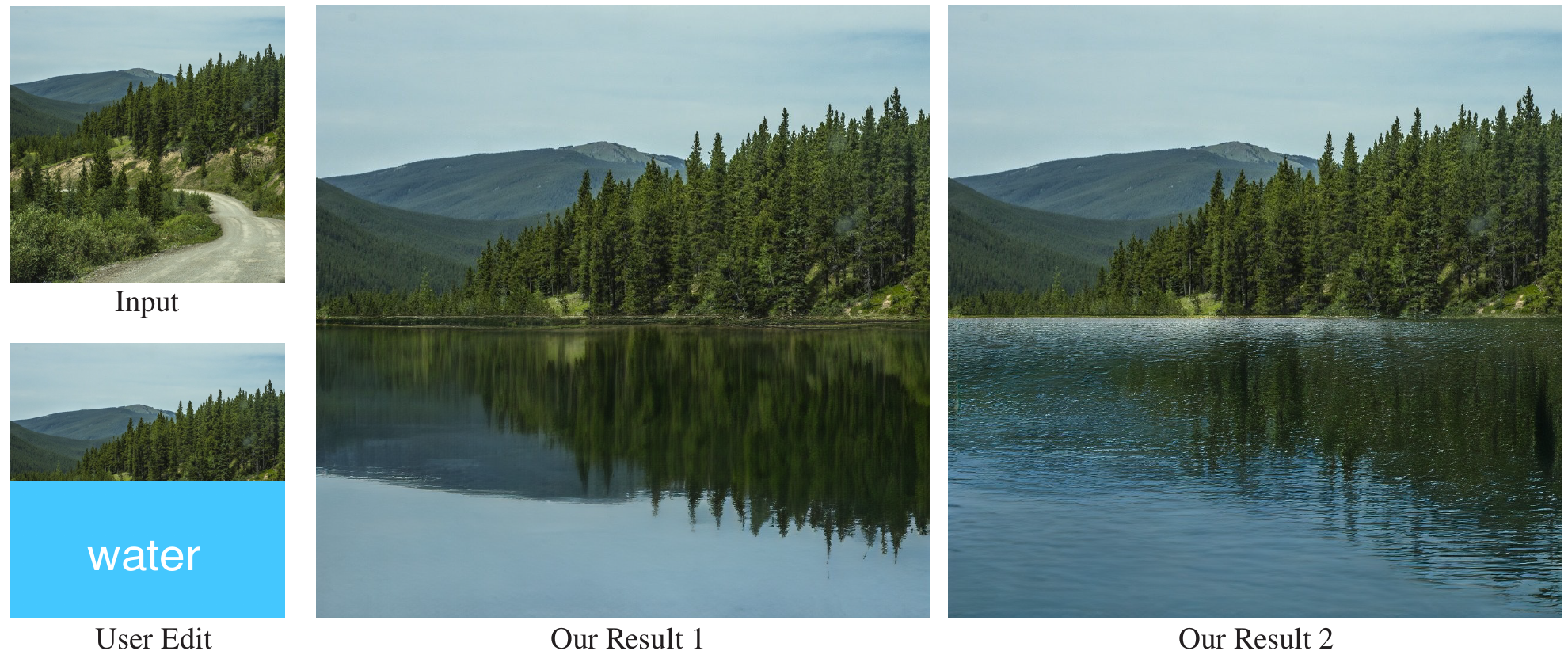}
\caption{
\rev{ASSET allows users to create diverse editing results by specifying a region and a new label on the input image. Our efficient transformer captures long-range dependencies in the image, such as the detailed reflection of the trees on the water, even at high resolutions ($1024\times1024$ pixels in this example).}
}
\vspace{2mm}
\label{fig:teaser}
\end{teaserfigure}
\maketitle

\section{Introduction}
\label{sec:intro}

Semantic image editing allows users to easily edit a given image by modifying a corresponding segmentation map.
Ideally, one could change the outline of existing regions, or even freely add or remove regions.
However, to obtain realistic and consistent results, an effective system needs to consider global context, from across the full image.
For example, consider the example in Figure~\ref{fig:teaser}. To properly hallucinate a reflection in the water on the bottom of the image, the model should consider the content from the very top.
Traditional CNN based approaches \cite{isola2017image,chen2017photographic,park2019semantic,ntavelis2020sesame,zhu2020sean} rely entirely on convolutional layers which have difficulty modeling such long-range dependencies \cite{wang2018non}.

Transformers are well equipped to handle these long-range dependencies through their attention mechanism allowing them to focus on distant image areas at each sampling step.
However, the heavy computational cost for using attention, which usually increases quadratically with the input size, makes it infeasible to use standard transformers for high-resolution image editing.
One way to address this is to use a sliding-window approach \cite{esser2021taming}, in which the transformer only attends to a small area around the currently sampled token, thereby reducing the computational cost to a fixed budget.
While this approach enables the synthesis of high-resolution images, it forgoes the benefit of modeling long-range dependencies.
This leads to inconsistencies when edits are dependent on image regions that are far away in pixel space. Is it possible to retain the ability to model long-range dependencies, while not paying prohibitive computational costs at high resolution?

We introduce a novel attention mechanism, called \emph{Sparsified Guided Attention} (SGA), to facilitate long-range image consistency at high resolutions.
While the sliding window approach is limited to local contexts, SGA can attend to far contexts that are relevant for the current sampling location. The core idea is to efficiently determine a small list of relevant locations that are worth attending to, and compute the attention map only over these locations. 
To achieve this, we use a guiding transformer that operates at the downsampled version of the input image and performs the same edit, but enjoys the full self-attention map thanks to the reduced input size. 
Based on the guiding transformer's attention map, we rank the importance of different areas of the image, and have the high resolution transformer attend only to the top-$K$ most important image regions.
In practice, our SGA leads to a large reduction in computational cost due to the obtained sparse attention matrix.
Compared to other approaches, we obtain more realistic and consistent edits while still achieving high diversity in our outputs.

Our model takes as input a quantized representation of the image and its edited segmentation map, both obtained through a modified VQGAN encoder \cite{esser2021taming}.
We then mask out all image tokens in the image representation corresponding to the edited area and replace those tokens with a specific \code{[MASK]} token.
Our transformer then samples new image tokens at the edited areas, conditioned on the original (masked) image and the edited segmentation map.
Finally, a VQGAN decoder is used to decode the image tokens into the final RGB image.
As the edited tokens are sampled autoregressively based on a likelihood-based model, we can sample a diverse set of image outputs, all of which are consistent with the overall image characteristics.

\paragraph{Contributions.} We propose a transformer-based model that outputs realistic and diverse edits specified through modified segmentation maps.
We introduce \emph{Sparsified Guided Attention} (SGA), which allows the transformer to only attend to the most important image locations, leading to sparse attention matrices and reduced computational cost.
Our model achieves diverse, realistic, and consistent image edits even at $1024 \times 1024$ resolution.

\section{Related Work}
\label{sec:related}
\paragraph{CNN-based image editing.}
CNN-based methods have achieved impressive results by enabling users to move bounding boxes containing objects \cite{hong2018learning,hinz2019generating}, modifying scene representations \cite{dhamo2020semantic,su2021fully}, or following textual instructions \cite{nam2018text,cheng2020sequential,patashnik2021styleclip}.
Other approaches enable user guides with edges or color information \cite{jo2019sc, liu2021deflocnet} or perform simple inpainting, typically without user guidance \cite{liu2018image, yu2018generative, yu2019free, yang2017high, suvorov2021resolution, liu2021pd}.
Exemplar-based image translation methods \cite{zhang2020cross, zhou2021cocosnet, zheng2020semantic} can synthesize images from semantic maps, but they cannot hallucinate new content that does not exist in the exemplar image.
Other approaches fine-tune or train a generator for a specific image to perform editing on that single image \cite{bau2020semantic, shaham2019singan,hinz2021improved,vinker2021image}. However, these models need to be adapted for each new image.
More similarly to us, other methods allow for direct editing via segmentation maps \cite{gu2019mask,lee2020maskgan,ntavelis2020sesame,ling2021editgan}. However, these approaches can only generate a single output for a given edit. In addition, previous CNN-based approaches prioritize local interactions between image pixels for image synthesis due to their inductive bias. They also fail to  effectively capture long-range interactions between image regions necessary for realistic image synthesis. Our approach is based on a transformer that is able to effectively capture such interactions and also allows the synthesis of diverse results for each edit. 

\paragraph{Transformers for image synthesis.}
To apply transformers for image synthesis, they are trained on discrete sequences of image elements. Some models first learn a discrete image region representation \cite{ramesh2021zero,esser2021taming}, whereas other approaches work directly on pixels \cite{parmar2018image,child2019generating,chen2020generative,jiang2021transgan}.
However, most of them model images in a row-major format, and thus cannot capture bidirectional context, leading to inconsistent editing results.
PixelTransformer \cite{tulsiani2021pixeltransformer}, iLAT \cite{cao2021image}, and ImageBART \cite{esser2021imagebart} add bidirectional context to their transformer models but do not support editing via segmentation maps. More importantly, due to their quadratic complexity in the number of image tokens, these methods are trained on small image resolutions of $256 \times 256$ pixels.
Alternatively, some approaches model the image directly at a low resolution (e.g., $32\times 32$) and then use a deterministic upsampling network \cite{wan2021high,yu2021diverse}.
In this case, fine-grained edits are difficult to achieve due to the small resolution at which the images are modeled. For high-resolution image synthesis, \cite{esser2021taming, esser2021imagebart}  proposed transformers with attention constrained on sliding windows.
However, this hampers long-range interactions and, as a result, they often generate artifacts and inconsistent image edits. In contrast, our work incorporates a novel sparsified attention mechanism that can capture such interactions without compromising the synthesized image plausibility.

\paragraph{Efficient transformers.}
Much work has been invested in reducing the computational cost of the transformer's attention mechanism \cite{tay2020efficient}.
Broadly speaking, there are two ways to achieve this.
One way is to reduce the computational cost of the attention mechanism directly, e.g., by approximating full attention through mechanisms where the computation cost grows linearly with the input length \cite{kitaev2020reformer,wang2020linformer}.
Alternatively, several works explore reducing the cost by replacing full attention with a sparse variant \cite{beltagy2020longformer,zaheer2020big}.
A few recent vision transformers reduce the computational complexity by spatially reducing attention \cite{liu2021swin, wang2021pyramid, zhang2021multi,yang2021focal}.
However, these methods use encoder-only transformers for feature extraction and do not support autoregressive image generation.
Our method is inspired by BigBird's sparse attention mechanism for long-document NLP tasks \cite{zaheer2020big}. BigBird achieves high efficiency by using a random sparse attention map over blocks of tokens.
However, when applying the random attention mechanism of BigBird to our task it fails to capture correct context for a given edit. Instead of randomly choosing tokens, our approach picks the most relevant tokens for attention at each spatial location.

\paragraph{Image synthesis with a guidance image.} Synthesizing high resolution outputs is challenging in terms of both quality and computational cost. The idea of utilizing a high-resolution guide to upsample a low-resolution output has been explored in computer graphics~\cite{kopf2007joint,chen2016bilateral}. In particular, constructing a high-resolution depth map from coarse sensor data, guided by an RGB image has been extensively investigated~\cite{yang2007spatialdepth,park2011upsampling,Ferstl_2013_ICCV,liu2013joint,Yang2014ColorGuidedDR}. More recently, learning-based approaches were developed for similar tasks, by posing it as an image-to-image translation problem~\cite{lutio2019guided}, fusing the guidance and low-res information at multiple scales~\cite{hui16msgnet}, or transferring the mapping learned at low resolution to high resolution~\cite{shocher2018zero}. While these works primarily aim at leveraging high-resolution information in the input as a guide, our application must synthesize information from a flat input. In fact, our guide is a low-resolution version of the same image. Relatedly, \cite{shaham2021spatially} use a low-resolution network to predict parameters of a lightweight high-resolution network, for the purpose of fast image translation, using a convolutional network architecture.

\section{Method}

\begin{figure}[t!]
  \centering
  \includegraphics[width=0.97\linewidth]{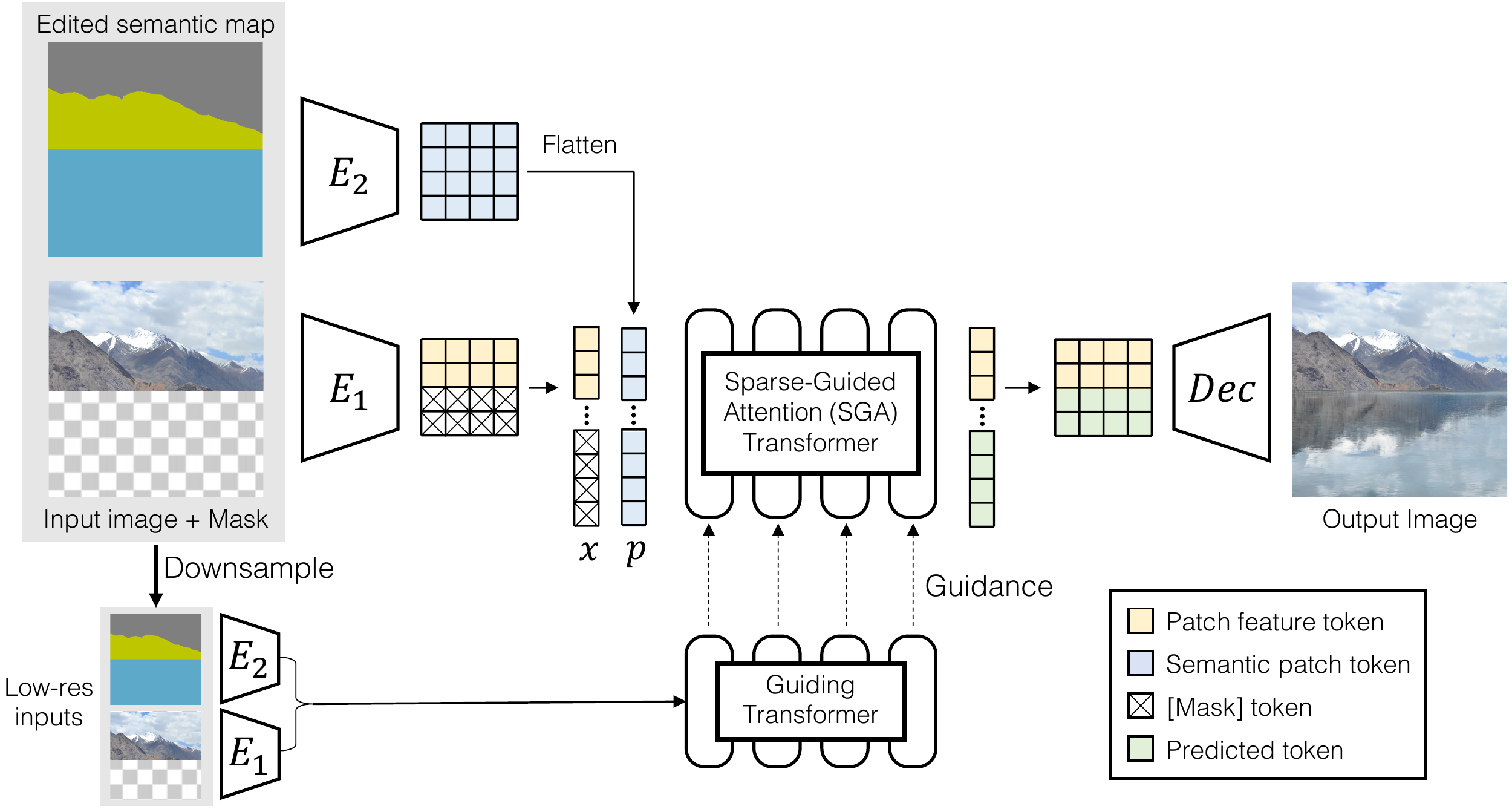}
  \caption{\rev{Overview of our semantic image editing model. Our transformer model operates in the codebook space using the encoder $E_1$. To incorporate user edits the tokens inside the edited region are masked and are augmented with a semantic encoder $E_2$. The mask tokens are filled in by our SGA-Transformer Encoder-Decoder network, whose sparse attention mechanism is guided by the Guiding Transformer that computes the full attention on downsampled inputs. Finally, the generated tokens are decoded into the output image via $Dec$.}}
  \label{fig:architecture}
\end{figure}

\paragraph{Overview.}
Our method synthesizes images guided by user input in the form of an edited label map (``semantic map'') of an input image.
More specifically, given an  RGB\ image  and its corresponding label map, the user paints some desired changes on the label map, e.g., replace mountain regions with water (Figure~\ref{fig:architecture}).
Since there exist several possible output images reflecting the input edits, our method generates a diverse set of outputs allowing the user to select the most preferable one.
Moreover, our method generates high-resolution images of up to $1024 \times 1024$ resolution. 

Our architecture is shown in Figure~\ref{fig:architecture}.
Inspired by recent approaches \cite{esser2021taming} we represent images and label maps as a spatial collection of quantized codebook entries (Section~\ref{sec:image_encoder}).
These codebook entries are processed by a transformer model which aims to update the codebook entries of the edited areas in an autoregressive manner (Section~\ref{sec:transformer}).
All codebook entries are subsequently decoded to the output set of images (Section~\ref{sec:image_decoder}).
A crucial component of the transformer is its attention mechanism, which enables long-range interaction between different parts of the image such that the synthesized output is coherent as a whole.
E.g., if a lake is generated by the semantic edits it must also capture any reflections of landscape (Figure~\ref{fig:teaser}).
One complication is that the quadratic complexity of the traditional attention mechanism leads to a large time and memory cost for high-resolution images.
The key idea of our method is to compute the full attention at lower resolution first, and then use that as guidance for sparsifying the attention at full resolution (Section~\ref{sec:transformer}).
This approach allows us to model long-range dependencies even at high resolutions, resulting in more coherent and plausible output images compared to existing approaches that constrain attention within sliding windows \cite{esser2021taming} or alternative attention models \cite{zaheer2020big}. 

\begin{figure*}[t!]
  \centering
  \includegraphics[width=\linewidth]{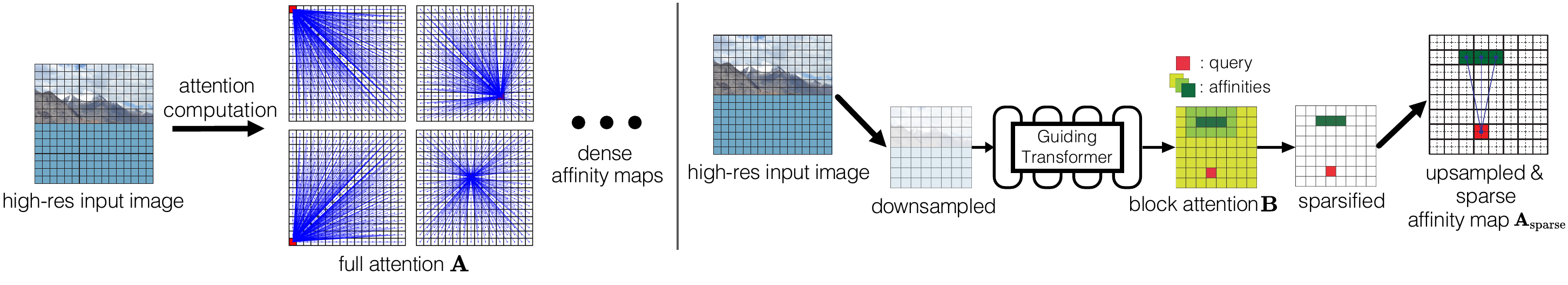}
  \caption{\rev{Details of our Sparsified Guided Attention (SGA, right) compared to full attention (left). We downsample the input and the user-edited semantic layout and use a Guiding Transformer to obtain the full attention map, which identifies the most important attention locations for each sampling step. By keeping only the locations with high attention weight, we construct the high-resolution, sparse attention map for the SGA transformer.}}
\label{fig:attention}
\end{figure*}

\subsection{Image encoder}
\label{sec:encoder}
\label{sec:image_encoder}

The input RGB image $\bX$ of size $H_\text{im} \times W_\text{im} \times 3$ is processed by a convolutional encoder resulting in a feature map $\bF$ of size $\frac{H_\text{im}}{16} \times \frac{W_\text{im}}{16} \times d$.
We also create a $H_\text{im} \times W_\text{im}$ binary mask indicating image regions that must be replaced according to the semantic map edits.
Masked image regions should not affect features produced for unmasked regions, e.g., information about the edited area of Figure~\ref{fig:architecture} should not ``leak'' into the feature map of the unmasked area.
To avoid information leakage, we employ partial convolutions \cite{liu2018image} and region normalization \cite{yu2020region} in our encoder while processing the unmasked regions.
The feature map $\bF$ is subsequently quantized following VQGAN \cite{esser2021taming} with the help of a learned codebook $\Z$, i.e.,  each feature map entry $\bff_{i,j}$ at position $(i,j)$ in $\bF$ is mapped to the closest codebook entry \mbox{$\hat\bff_{i,j}=\argmin_{\bz_{\kappa} \in \Z} ||\bff_{i,j} - \bz_{\kappa}||$}, where $\{\bz_{\kappa}\}_{{\kappa}=1}^{|\Z|}$ are  codebook entries with dimensionality $d$.
The codebook indices of the edited regions, as indicated by the binary mask, are replaced with a special \code{[MASK]} token (Figure \ref{fig:architecture}).
We use a second encoder with regular convolutions to obtain a feature representation of the edited semantic map $\bP$ which is subsequently quantized in the same way as the RGB image, resulting in codebook entries $\hat\bg_{i,j}$.

\subsection{Autoregressive transformer}
\label{sec:transformer}
Our transformer follows a sequence-to-sequence architecture inspired by \cite{vaswani2017attention} which consists of a bidirectional encoder and an autoregressive decoder, both of which are equipped with our novel sparsified attention mechanism.
The transformer encoder captures bi-directional context of the image, which is used by the transformer decoder to generate new codebook indices autoregressively.

\paragraph{\rev{Traditional Dense Attention.}}
Traditional attention transforms each embedding linearly into a learned query, key, and value representation $\bQ,\bK,\bV$ of size $L \times d$, where $L=H_\text{feat}W_\text{feat}=\frac{H_\text{im}}{16}\frac{W_\text{im}}{16}$ is the length of the flattened codebook indices in our case \cite{vaswani2017attention}.
The output embedding is then computed as \mbox{$\mathrm{softmax}(\bA/\sqrt{d})\bV$}, where attention $\bA = \bQ \bK^T\in \mathds{R}^{L\times L}$.
The advantage of attention is that it allows for interactions across all positions in the sequence, i.e., in our case the whole encoded image, as illustrated in Figure~\ref{fig:attention} {\it (left)}.
The disadvantage is computation of the attention matrix $\bA$ has quadratic time and memory complexity in terms of sequence length: $\mathcal{O} (L^2) = \mathcal{O} (H_\text{feat}^2 W_\text{feat}^2)$.
For an input image with resolution $1024 \times 1024$, the sequence has length $L=4096$, and practically the cost of performing the above matrix multiplication becomes prohibitively high, as discussed by several other works \cite{tay2020efficient}. 
One simple way to reduce the computational cost is to use a sliding window approach \cite{esser2021taming}, i.e., crop a fixed sized patch around the currently sampled token and feed it into the transformer.
However, this introduces a bias towards preferring interactions only within local regions of the image, missing other useful longer-range interactions.

\paragraph{Sparsified  Guided Attention (SGA)} We propose an efficient sparsification strategy, without sliding windows. The key idea is to first compute coarse full attention with downsampled images, determine which locations are worth attending to, and then use it to avoid computing most entries of the attention matrix $\bA$. 

To do this, we first proceed by downsampling the original input image and semantic map to $256 \times 256$ resolution. We further encode them to obtain a feature map of size $16 \times 16$. At this resolution, computing full attention is fast, because the sequence length of codebook indices is only $16^2 = 256$. Then we employ a {\it guiding transformer}, which has the same architecture as the main transformer but is trained at the low resolution, to calculate the dense attention matrix $\bA_{\text{low}} \in \mathds{R}^{256\times 256}$.

Then we leverage $\bA_{\text{low}}$ to construct a block attention matrix $\bB \in \mathds{R}^{L\times L}$ at the original feature resolution that will guide the sparsification. To do this, we divide the original feature map into non-overlapping blocks, as illustrated in Figure~\ref{fig:attention} {\it (right)} for an $8 \times 8$ grid of blocks. For each block, we find the corresponding locations in $\bA_{\text{low}}$ and average their affinity values. Note that the matrix $\bB$ essentially consists of blocks, each of which is populated with a single affinity value.

Then we construct the sparse attention matrix $\bA_{\text{sparse}}$ by considering only the attention weights that are likely important in approximating the true attention. To this end, we keep the attention if the corresponding affinity is high in the block attention matrix $\bB$. In other words, we argue that the selection of sparse attention pairs can be reliably guided by the dense attention evaluated at lower resolution. In addition, following the proposition in the context of NLP models \cite{zaheer2020big}, we always compute attention within the current and adjacent blocks, no matter their affinity values.

\begin{equation}
  \bA_{\text{sparse}}(r, t)=\begin{cases}
    \bA(r, t), & \text{if $t \in \N(r)$ or  $t \in \K(r)$}.\\
    -\infty, & \text{otherwise},
  \end{cases}
\end{equation}

\noindent where $\N(r)$ contains the entries of the neighborhood blocks of $r$, and $\K(r)$ contains the entries of the blocks with the top-$K$ highest affinities outside the neighborhood.

In our experiments, we set $K=3$, resulting in the sparsity ratio ${<}10\%$, and significantly reduce the computational cost of attention.

\paragraph{Transformer encoder.} 
The input to our transformer encoder is a sequence of embeddings jointly representing the masked image codebook indices $\bx = \{ x_l \}_{l=1}^{L}$ and semantic codebook indices $\bp = \{ p_l \}_{l=1}^{L}$ produced by the image encoders (flattened using row-major format), and position information of each index in the corresponding sequence. Note that here, the transformers are operating both at full-resolution and low-resolution.
Specifically, for each position in the sequence, three $d$-dimensional learned embeddings are produced: (i) an image embedding $E_\text{im}(x_l)$ representing the token $x_l$ at position $l$ in our sequence and in turn the corresponding RGB image region, (ii) an embedding $E_\text{map}(p_l)$ of the semantic token $p_l$ at the same position, and finally (iii) a positional embedding $E_\text{pos}(l)$ for that position $l$.
The summation of token embeddings and positional embeddings follows other popular transformers \cite{vaswani2017attention, dosovitskiy2020image}:
\begin{equation}
    \be_l = E_\text{im}(x_l) + E_\text{map}(p_l) + E_\text{pos}(l)
\end{equation}
The sequence of embeddings $\{\be_l\}_{l=1}^{L}$ is fed to the first transformer encoder layer and is transformed to a continuous representation by the stack of transformer encoder layers.
To preserve the position information of each index at subsequent layers, the position encoding generator (PEG) is placed before each encoder layer \cite{chu2021conditional, chu2021twins} (more details in the appendix).

\paragraph{Transformer decoder.} 
The decoder predicts codebook indices for the edited region with the help of the global context obtained through the transformer encoder.
Similar to BART \cite{lewis2019bart}, the autoregressive generation starts by pre-pending a special index (token) \code{[START]} to the decoder input.
At each step, the decoder predicts a distribution over codebook indices from our dictionary $\Z$ learned in our image encoder (Section~\ref{sec:image_encoder}).
Specifically, the decoder predicts $p(\X_{l} | \{\chi_{<l}\} )$, where $\X_l$ is a categorical random variable representing a codebook index to be generated at position $l$ in the sequence and $\{\chi_{<l}\}$ are all indices of the previous steps.
We note that the tokens corresponding to unmasked image regions (i.e., image regions to be preserved) are set to the original image codebook indices.
We predict the distributions only for positions corresponding to the edited image regions.

To predict the output distribution at each step, the decoder first takes as input a learned embedding $D_\text{im}(x_l)$ representing the input token $x_l$, and a learned positional embedding $D_\text{pos}(l)$ for that position $l$.
It sums the two embeddings $\bd_l = D_\text{im}(x_l) + D_\text{pos}(l)$, then passes $\bd_l$ into a self-attention layer (attention between generated tokens) and a cross-attention layer (attention between generated tokens and encoder output features).
For both self-attention and cross-attention we make use of the sparsified guided attention mechanism.
We also note that the self-attention in the decoder layer is modified to prevent tokens from attending to subsequent positions. 
For more details about the architecture of our decoder, we refer to the appendix.
Based on the predicted distribution of codebook indices, we use top-k sampling \cite{holtzman2020degeneration,esser2021taming} ($k=100$ in our experiments) to create multiple candidate output sequences, each of which can be mapped to a new image by the image decoder. The generated images are ordered by the joint probability of the distributions predicted by the decoder.

\subsection{Image decoder}
\label{sec:image_decoder}

The image decoder takes as input the quantized feature map and decodes an RGB image of size $H_\text{im} \times W_\text{im} \times 3$ following VQGAN \cite{esser2021taming} (see appendix for architecture details).
Due to the quantization process, the reconstruction of the encoder-decoder pair is not perfect and leads to minor changes in the areas that are not edited.
To avoid this, we follow the same strategy as SESAME \cite{ntavelis2020sesame} and retain only the generated pixels in the masked regions while the rest of the image is retrieved from the original image.
To further decrease any small artifacts around the borders of the masked regions we apply Laplacian pyramid image blending as a final post-processing step.

\subsection{Training}
\label{sec:training} 
We randomly sample free-form masks following \cite{ntavelis2020sesame} and use the semantic information in the masked area as user edits. The image encoder, decoder, and transformer are trained in a supervised manner on training images which contain ground-truth for masked regions.
We first train our image encoders and decoders following VQGAN \cite{esser2021taming}. 
We then train our transformer architecture on images with $256 \times 256$ resolution using the original attention mechanism (full attention), which will be used as the \emph{guiding transformer}.
Following that, we switch to train  our SGA-transformer with the sparsified guided attention on high resolution, specifically, we initialize its weights from the previously trained guiding transformer and fine-tune it at $512 \times 512$ resolution again at $1024 \times 1024$ resolution.
In all cases, we use the same losses proposed in VQGAN \cite{esser2021imagebart}.
For more details about our model and training procedures please see the appendix.

\section{Results}
\rev{In this section, we present qualitative and quantitative evaluation for ASSET.}

\paragraph{Dataset.} 
We evaluate our ability to perform high-resolution semantic editing on the Flickr-Landscape dataset consisting of images crawled from the Landscape group on Flickr.
It contains $440K$ high-quality landscape images. We reserve $2000$ images for our testing set, while the rest are used for training.
Following \cite{ntavelis2020sesame}, we use $17$ semantic classes including mountain, clouds, water, and so on. 
To avoid expensive manual annotation, we use a pre-trained DeepLabV2 \cite{chen2017deeplab} to compute segmentation maps for all images.
\rev{We also use the ADE20K \cite{zhou2017scene} and COCO-Stuff \cite{caesar2018coco} datasets for 
 additional evaluation.}

\newlength{\itemheight}
{
\begin{figure*}[tb]
\centering
\def\fsh{\footnotesize}  
    
\renewcommand{\tabcolsep}{1px}  
\renewcommand{\arraystretch}{0.8}  
\setlength{\itemheight}{1.85cm}  

\begin{tabular}{ccp{1px}ccp{1px}ccp{1px}cp{1px}cp{1px}c}

     \includegraphics[height=\itemheight]{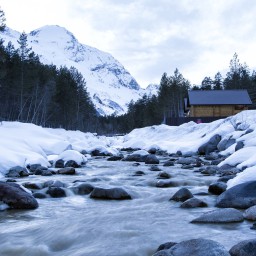} &
     \includegraphics[height=\itemheight]{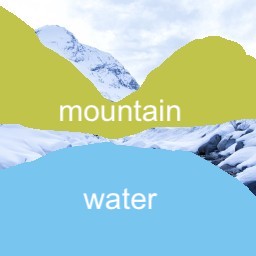} &&
     \includegraphics[height=\itemheight]{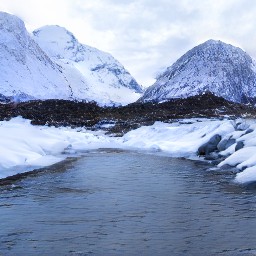} &
     \includegraphics[height=\itemheight]{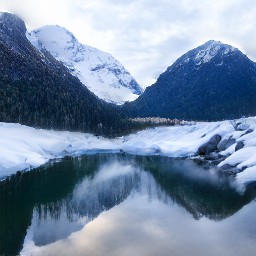} &&
     \includegraphics[height=\itemheight]{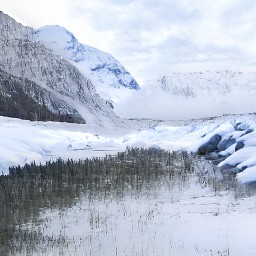} &
     \includegraphics[height=\itemheight]{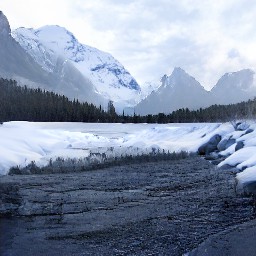} &&
     \includegraphics[height=\itemheight]{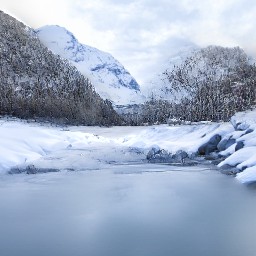} &&
     \includegraphics[height=\itemheight]{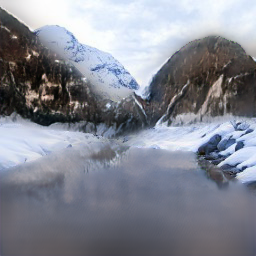} &&
     \includegraphics[height=\itemheight]{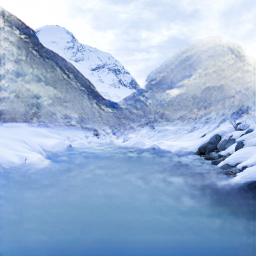} \\

     \includegraphics[height=\itemheight]{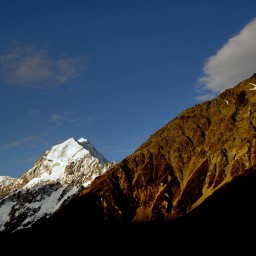} &
     \includegraphics[height=\itemheight]{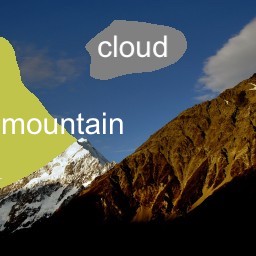} &&
     \includegraphics[height=\itemheight]{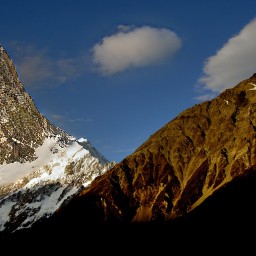} &
     \includegraphics[height=\itemheight]{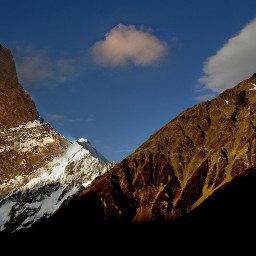} &&
     \includegraphics[height=\itemheight]{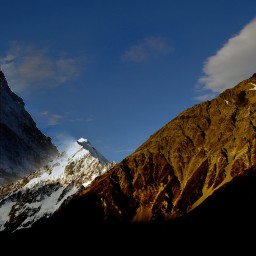} &
     \includegraphics[height=\itemheight]{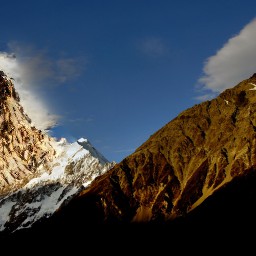} &&
     \includegraphics[height=\itemheight]{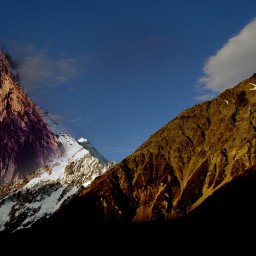} &&
     \includegraphics[height=\itemheight]{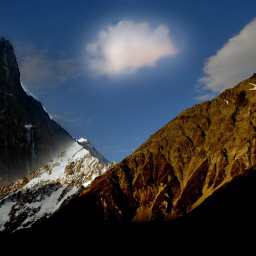} &&
     \includegraphics[height=\itemheight]{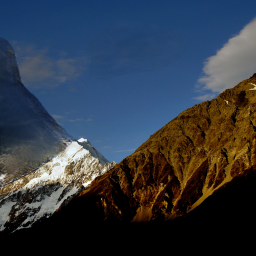} \\
     
     \fsh Input & \fsh Edits && \multicolumn{2}{c}{\fsh ASSET} && \multicolumn{2}{c}{\fsh Taming Transformers } && \fsh ImageBART && \fsh INADE && \fsh SESAME  \\
     
\end{tabular}
\caption{
\rev{Comparison of our approach with Taming Transformers \cite{esser2021taming}, ImageBART \cite{esser2021imagebart}, INADE \cite{tan2021diverse}, and SESAME \cite{ntavelis2020sesame}
on $256 \times 256$ images.}}
\label{fig:comparison_256}
\end{figure*}
}

\begin{table}[t]
\setlength{\tabcolsep}{2.5pt}
\centering
\caption{\rev{Quantitative evaluation on the Flickr-Landscape dataset at various resolutions.}}
\begin{tabular}{llcccccc}

\toprule
Res & Method & LPIPS $\downarrow$ & FID $\downarrow$ & SSIM $\uparrow$ & mIoU $\uparrow$ & accu $\uparrow$ & div $\uparrow$ \\
\midrule
\multirow{5}{*}{$256$} & \emph{INADE } & 0.233 & 11.2 & 0.826 & 48.6 & 59.1 & 0.145 \\
& \emph{SESAME} & 0.213  & 10.2 & 0.830 & 50.3 & 61.7 & 0.000 \\
& \emph{TT} &  0.201 & 10.4 & 0.839 & 46.1 & 58.3 & \textbf{0.187}  \\
& \emph{ImageBART} &  0.196 & 10.0 & 0.841 & 47.3 & 58.5 & 0.163  \\
& \emph{ASSET} & \textbf{0.187} & \textbf{9.2} & \textbf{0.846} & \textbf{51.5} & \textbf{63.0} & 0.151 \\
\midrule

\multirow{3}{*}{$512$} & \emph{TT} & 0.203 & 10.6 & 0.850 & 52.2 & 63.7 & \textbf{0.186}  \\
& \emph{ImageBART} &  0.199 & 10.4 & 0.851 & 52.4 & 63.3 & 0.168  \\
& \emph{ASSET} & \textbf{0.186} & \textbf{8.4} & \textbf{0.856} & \textbf{53.5} & \textbf{64.7} & 0.145 \\
\midrule

\multirow{3}{*}{$1024$} & \emph{TT} & 0.210 & 10.9 & 0.881 & 50.4 & 61.7 & \textbf{0.160}  \\
& \emph{ImageBART} &  0.201 & 10.4 & 0.880 & 50.8 & 62.1 & 0.139  \\
& \emph{ASSET} & \textbf{0.160} & \textbf{7.7} & \textbf{0.887} & \textbf{54.1} & \textbf{65.2} & 0.124 \\
\bottomrule
\end{tabular}
\label{table:comparison}
\end{table}

\paragraph{Evaluation metrics.}
To automatically generate test cases, we simulate user edits by masking out the pixels belonging to a random semantic class for each test image.
As explained in Section \ref{sec:transformer}, we can sample several output images and also rank them according to their probability. For our experiments, we sample $50$ images, and keep the top $10$ of them, as also done in other works \cite{liu2021pd, zheng2019pluralistic}. This results in $20$K generated images for our test set.
To evaluate the perceptual quality of our edits we compute the FID \cite{heusel2017gans}, LPIPS \cite{zhang2018unreasonable}, and SSIM metrics \cite{wang2004image}. For each ground-truth image of the test split, we evaluate these metrics against the synthesized image that achieves the best balance of them, as done in \cite{liu2021pd, zheng2019pluralistic}.
To evaluate how well the models adhere to the specified label map at the edited areas we also compare the mean Intersection over Union (mIoU) and the pixel-accuracy between the ground truth semantic map and the inferred one using the pretrained DeepLabV2 model~\cite{chen2017deeplab}.
Finally, to evaluate diversity, we utilize the LPIPS metric following \cite{zheng2019pluralistic, liu2021pd}.
The diversity score is calculated as the average LPIPS distance between $5$K pairs of images 
randomly sampled from all generated images, as also done in \cite{liu2021pd, zheng2019pluralistic}.
We also perform a user study to evaluate the perceptual quality of several models.

\paragraph{Baselines.}  
We compare our method with several semantic image editing baselines: SESAME \cite{ntavelis2020sesame}, INADE \cite{tan2021diverse}, Taming Transformers (TT) \cite{esser2021taming}, \rev{and ImageBART \cite{esser2021imagebart}}.
SESAME and INADE are based on convolutional networks and only support image resolutions of up to $256 \times 256$ pixels.
TT and ImageBART are based on Transformers, but use a sliding window approach at high resolution, in which the attention is only calculated on a local neighborhood for each sampling location.
While SESAME can only produce a single output, the other three methods can generate diverse outputs for a given edit. For a fair comparison, when generating image samples using TT or ImageBART, we also selected top-$10$ images out of $50$ sampled ones based on their probability. We selected top-$10$ images for INADE based on discriminator scores as done in \cite{liu2021pd, zheng2019pluralistic}.
For all baselines, we use the authors' implementation and train them on the same datasets as our method.

\cocoade

\begin{figure*}[t]
\centering
\includegraphics[width=0.93\textwidth]{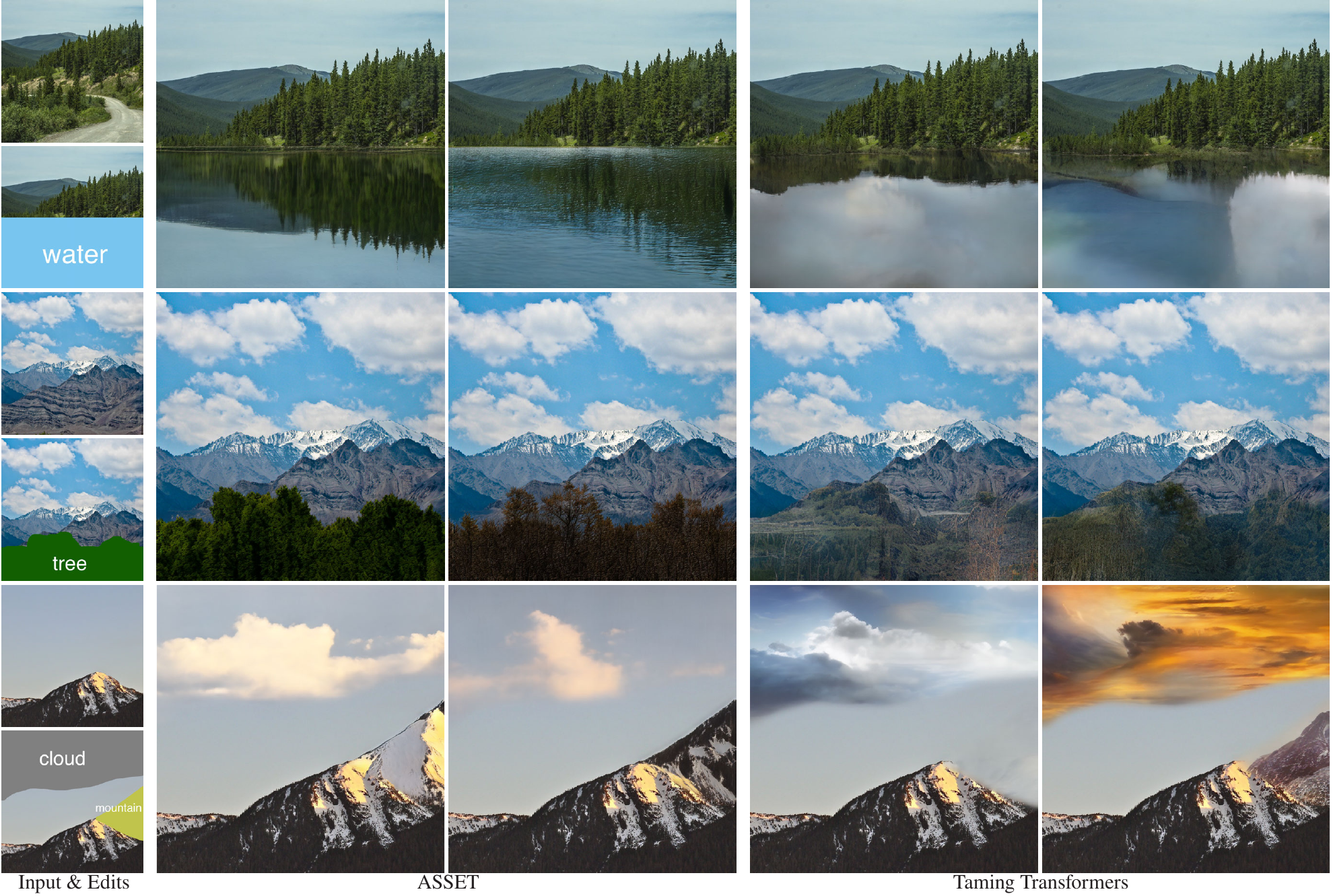}
\caption{
\rev{Comparison of ASSET with Taming Transformers \cite{esser2021taming} at $1024 \times 1024$ resolution.}}
\label{fig:comparison_1024_TT}
\end{figure*}

\begin{figure*}[t]
\centering
\includegraphics[width=0.93\textwidth]{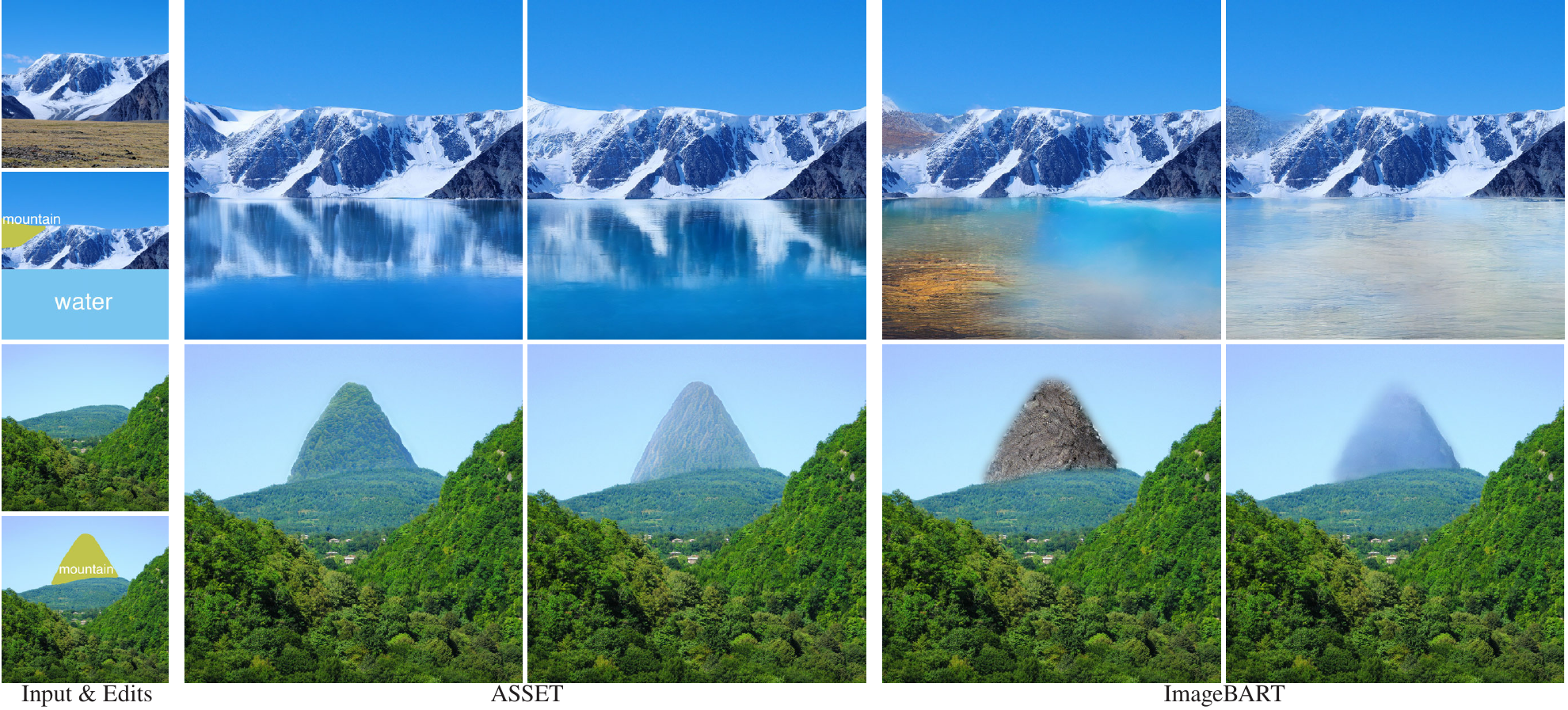}
\caption{
\rev{Comparison of ASSET with ImageBART
\cite{esser2021imagebart}
at $1024 \times 1024$ resolution.} }
\label{fig:comparison_1024_ImageBART}
\end{figure*}

{
\begin{figure*}[tb]
\centering
\def\fsh{\footnotesize}  
    
\renewcommand{\tabcolsep}{1px}  
\renewcommand{\arraystretch}{0.8}  
\setlength{\itemheight}{2.8cm}  

\begin{tabular}{cccccc}

     \includegraphics[height=\itemheight]{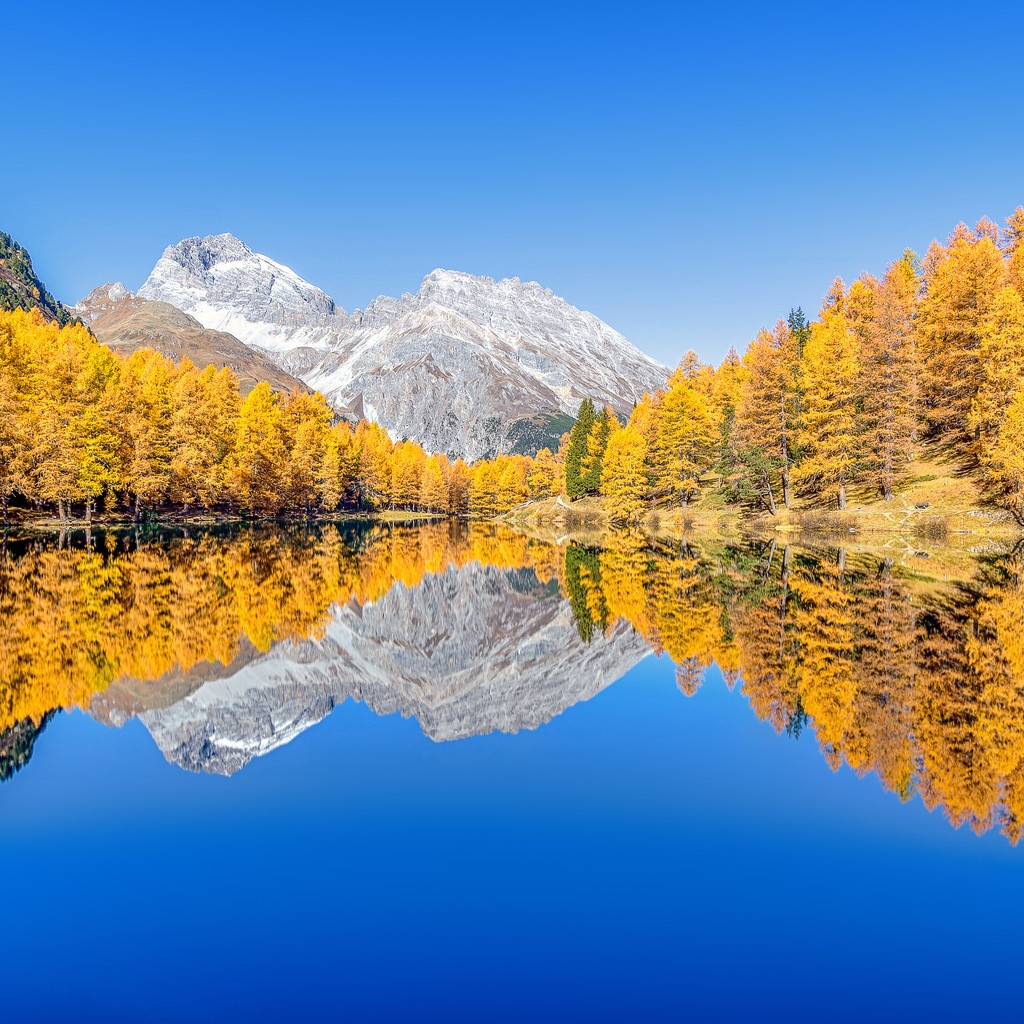} &
     \includegraphics[height=\itemheight]{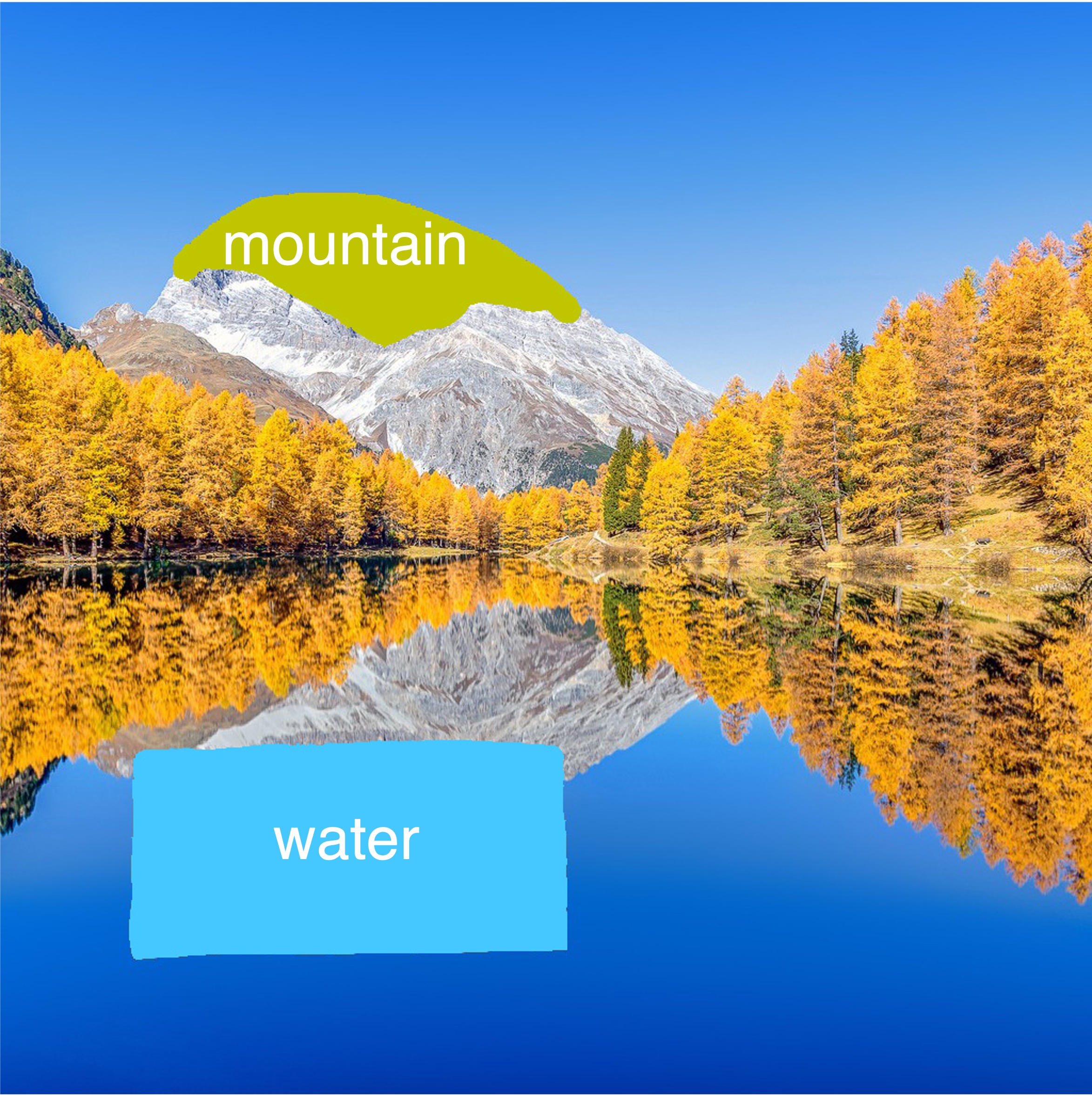} &
     \includegraphics[height=\itemheight]{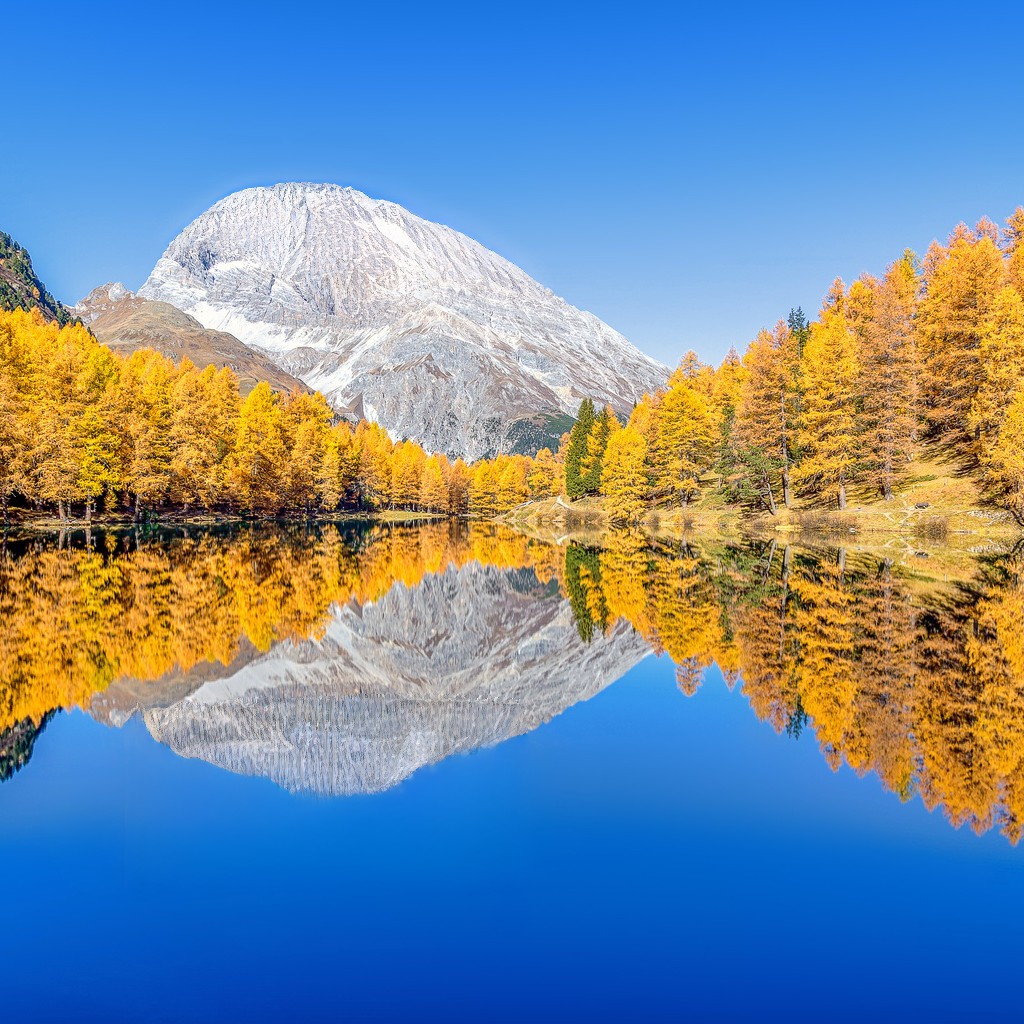} &
     \includegraphics[height=\itemheight]{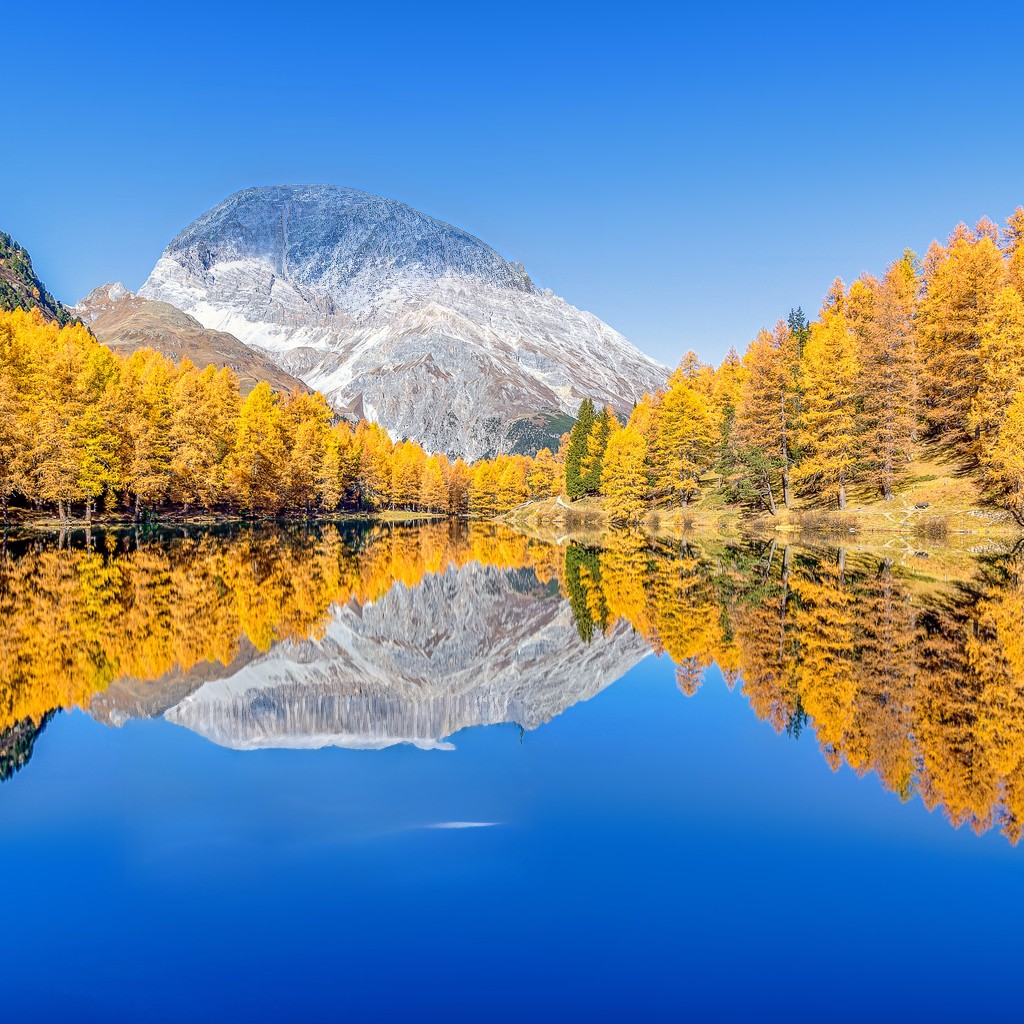} &
     \includegraphics[height=\itemheight]{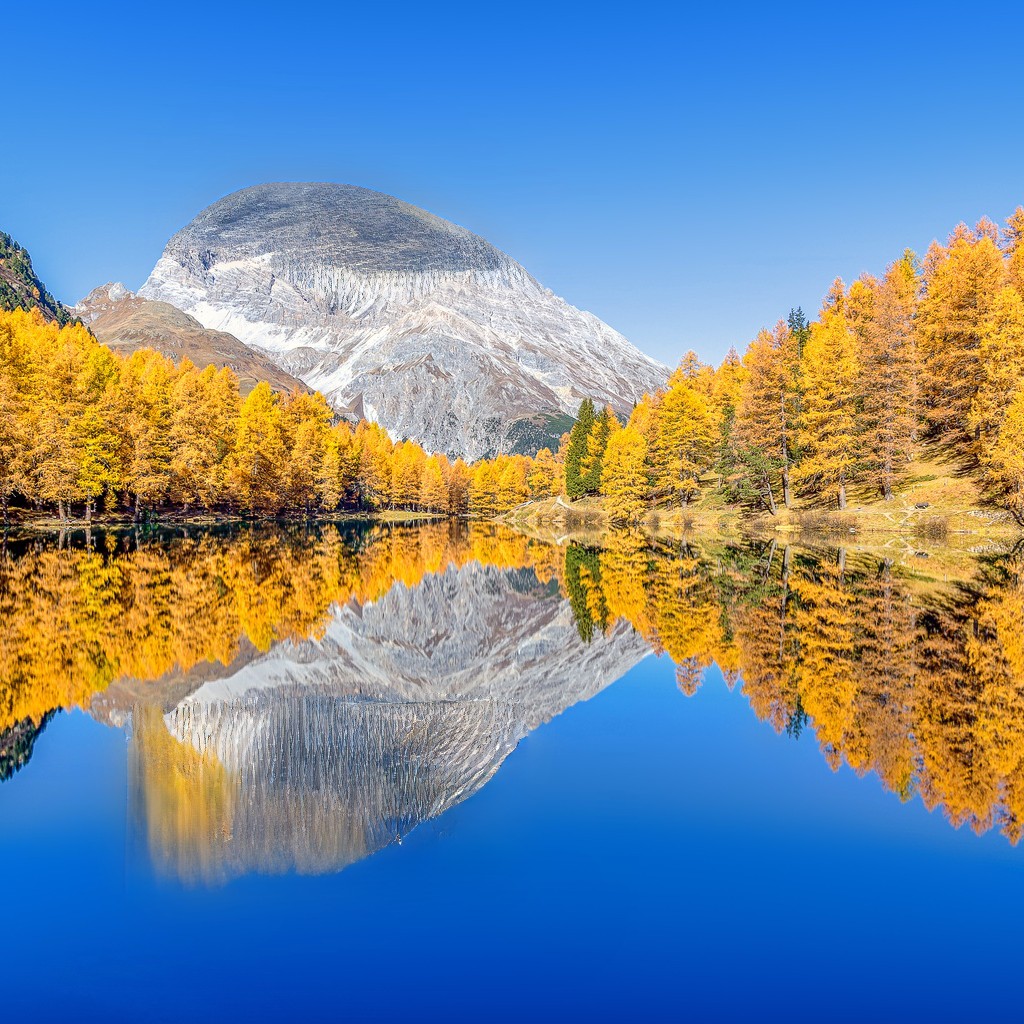} &
     \includegraphics[height=\itemheight]{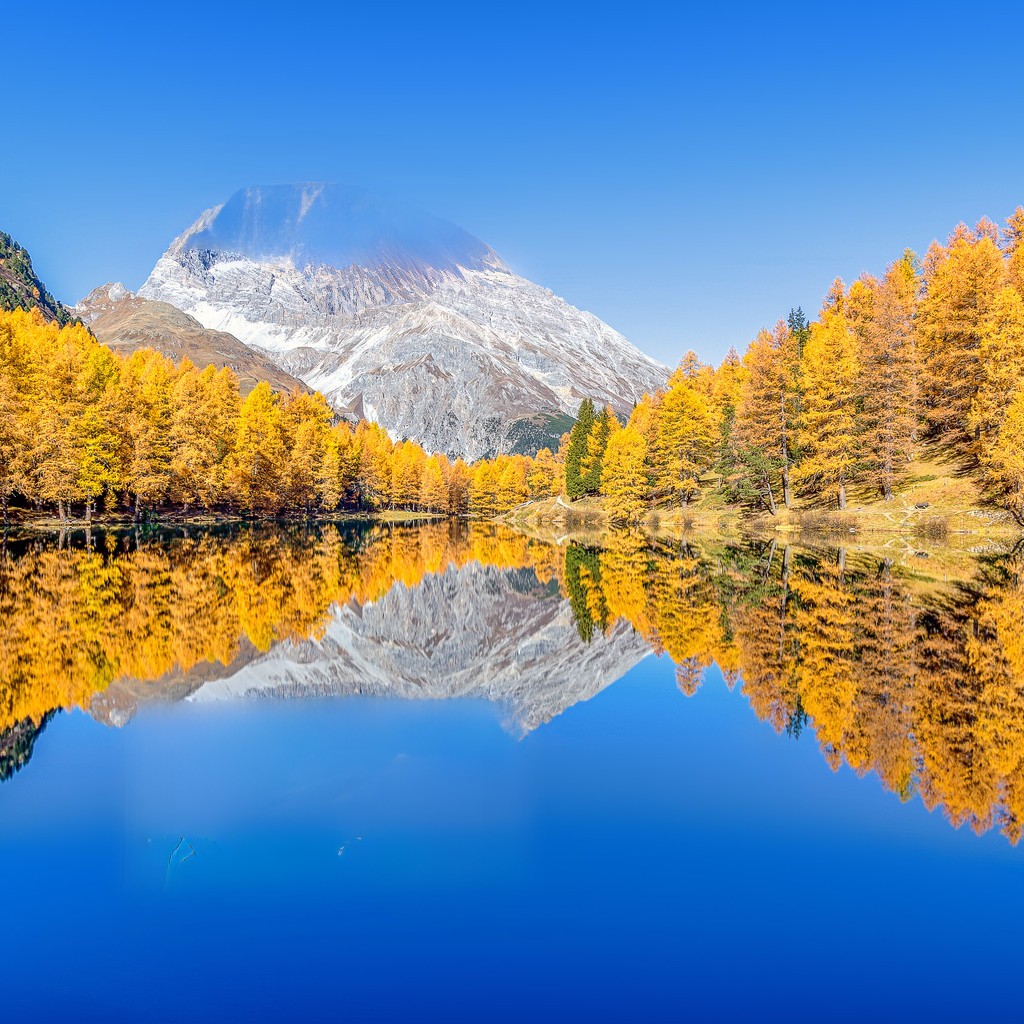} \\

     \includegraphics[height=\itemheight]{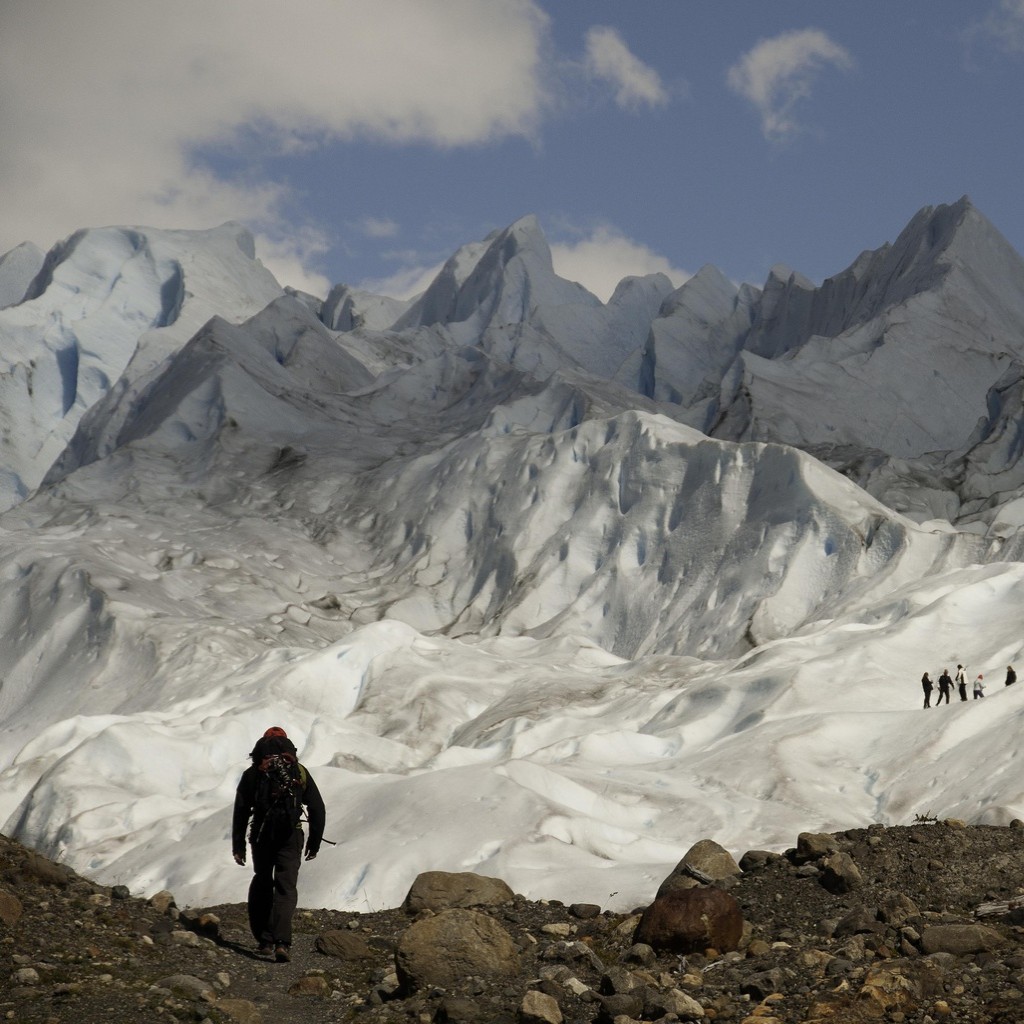} &
     \includegraphics[height=\itemheight]{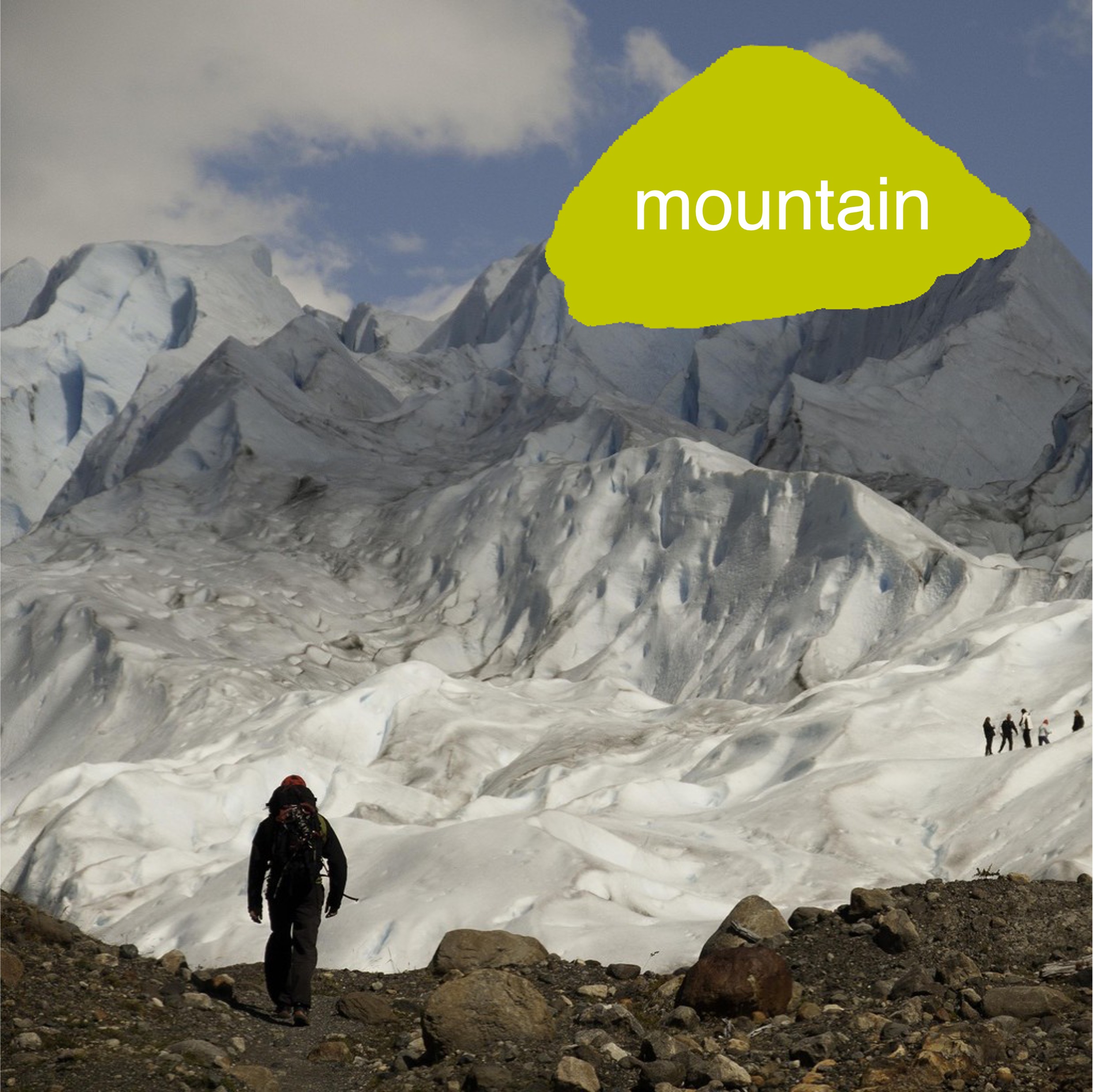} &
     \includegraphics[height=\itemheight]{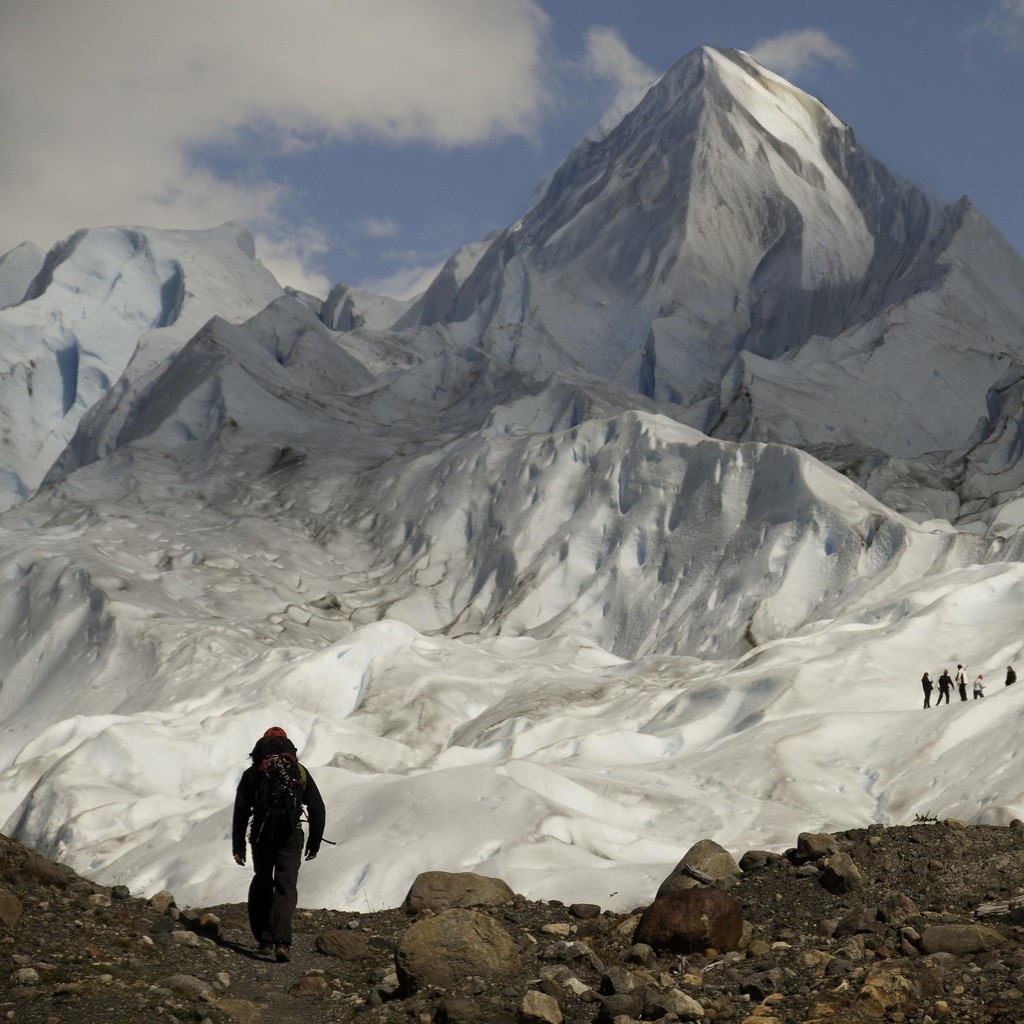} &
     \includegraphics[height=\itemheight]{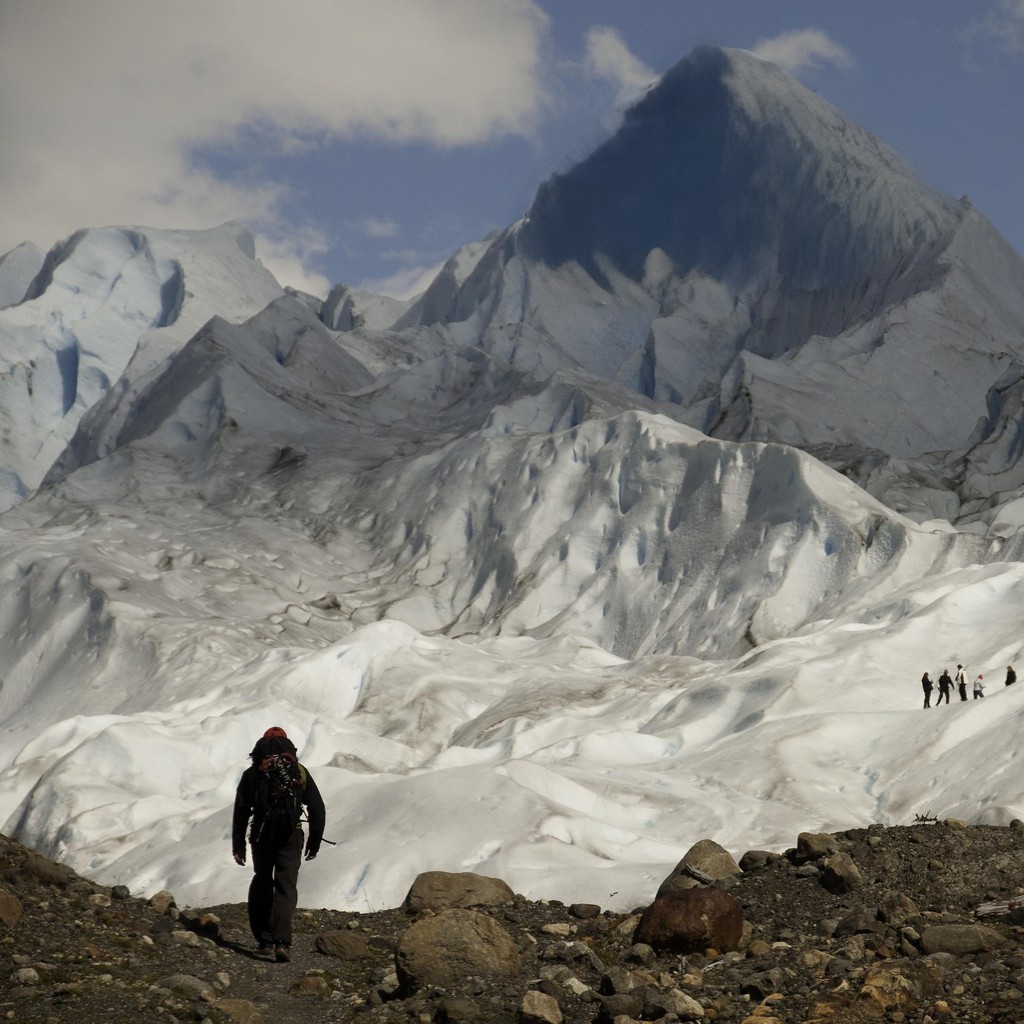} &
     \includegraphics[height=\itemheight]{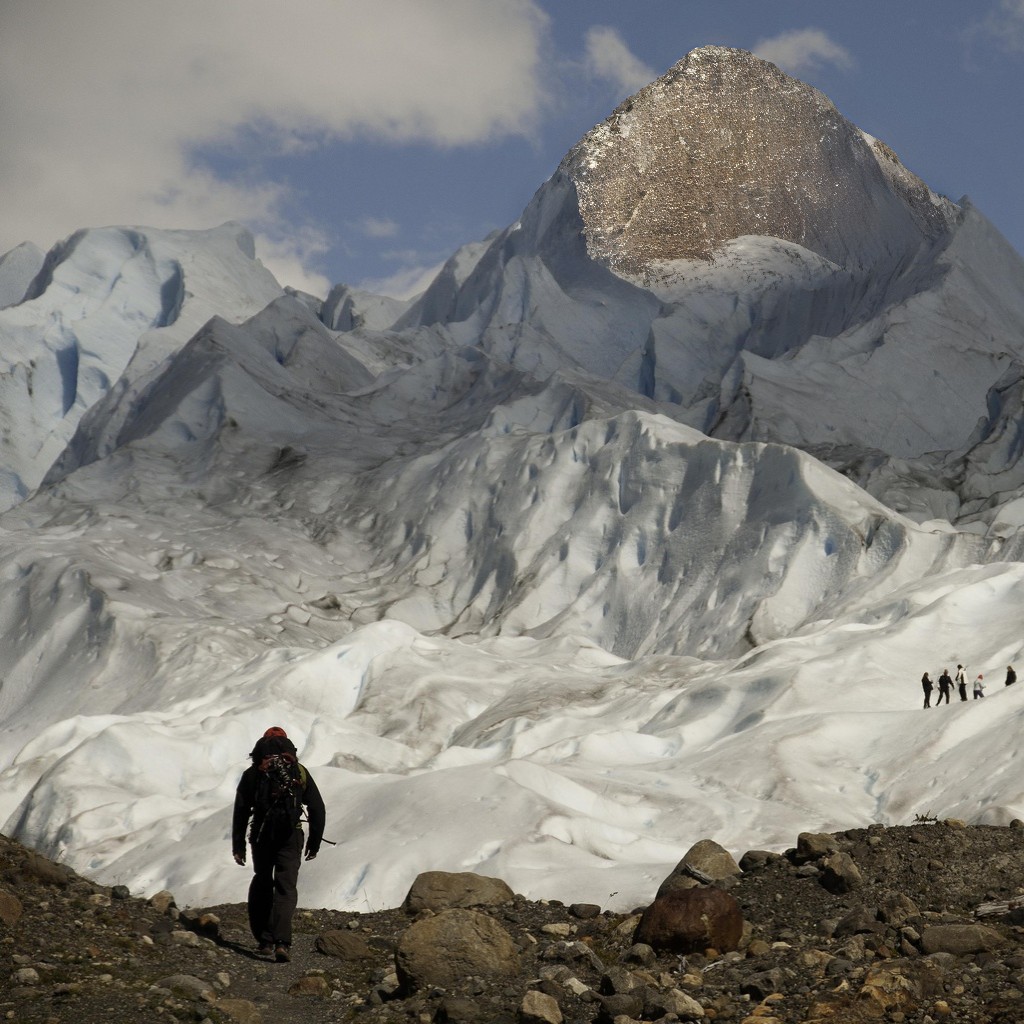} &
     \includegraphics[height=\itemheight]{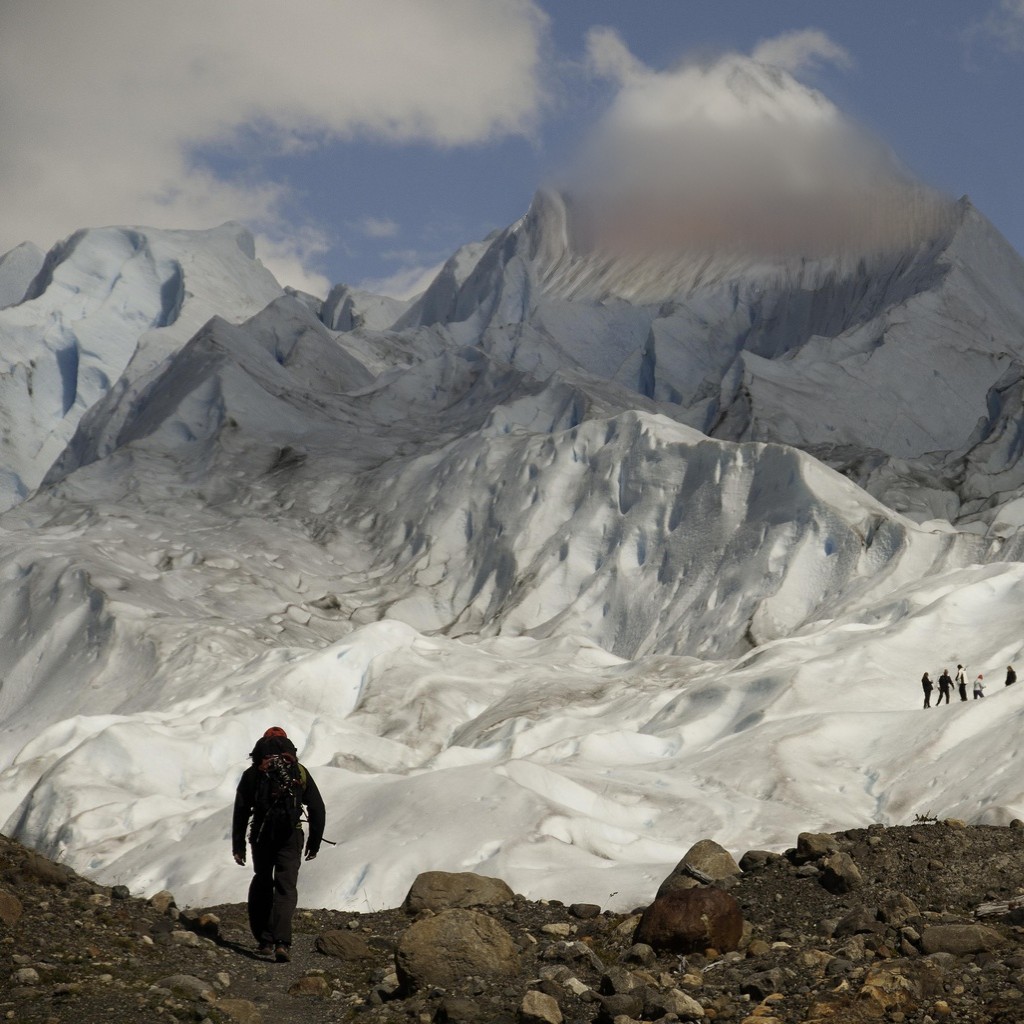} \\
     
     \fsh Input Image & \fsh User Edits & \fsh Ours & \fsh Random & \fsh Sliding & \fsh Local \\
\end{tabular}
\vspace{2mm}
\caption{Comparison showing the effects of using different kinds of attention at $1024 \times 1024$ resolution.}
\vspace{2mm}
\label{fig:ablation}
\end{figure*}
}

\paragraph{Quantitative evaluation.}
For multimodal editing tasks we aim to obtain outputs that are both diverse and consistent.
There is an inherent trade-off between diversity and consistency, as higher diversity can be achieved by sacrificing image consistency.
As such, we aim to achieve maximal diversity without generating inconsistent results.
For all models that can generate more than one solution for a given edit,  we choose the sample with the best balance of quantitative measures (out of the top 10 samples), as done in \cite{liu2021pd} and \cite{zheng2019pluralistic}.
Table~\ref{table:comparison} \rev{shows the comparison with competing models on the Flickr-Landscape dataset at different resolutions.}

Except for the diversity metric our model outperforms all competing methods on all resolutions.
While TT and ImageBART achieve a higher diversity than our model, we observe that this higher diversity comes at the cost of inconsistent images (see Figure~\ref{fig:comparison_1024_TT} and Figure~\ref{fig:comparison_1024_ImageBART}).
In contrast, our approach also achieves high diversity but shows much more consistent image outputs, both at lower and higher resolutions.
\rev{At low resolution ($256\times 256$), our method differs from TT by using a bidirectional transformer encoder to capture global context and
partial convolutions to
prevent information leakage from masked regions.}
As we increase the resolution ($512\times 512$ and higher), our approach continues to obtain consistent results, thanks to the Sparsified Guided Attention (SGA) that captures long-range dependencies.
In contrast,  the perceptual performance of TT and ImageBART decreases with increasing resolution, as the sliding window approach is unable to enforce consistency over long distances.

\rev{We also conduct experiments on the COCO-Stuff and ADE20K datasets.} Table~\ref{tab:cocoade} \rev{shows the comparison with TT at $512 \times 512$ resolution. Our model outperforms TT on both datasets.}

\begin{figure}[t!]
  \centering
  \includegraphics[width=0.97\linewidth]{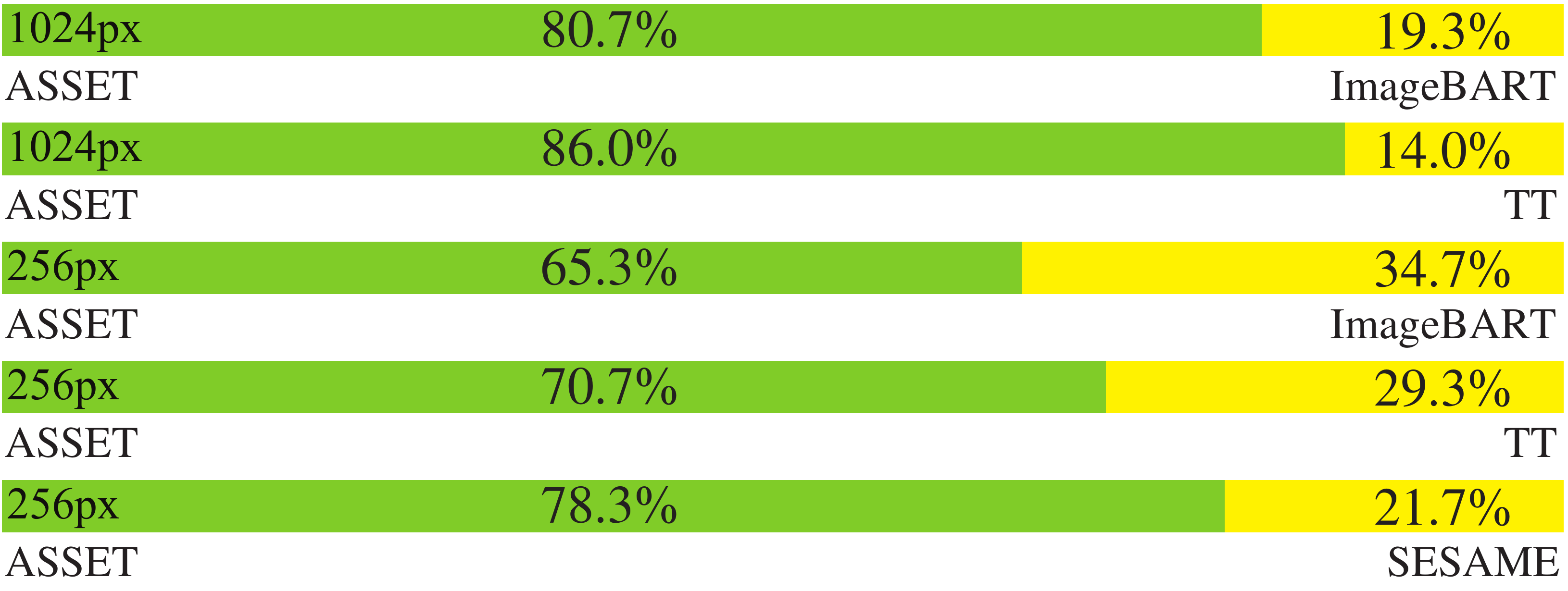}
  \caption{User study results. At both low and high resolution, our method ASSET is dominantly preferred over the baselines.}
  \label{fig:user_study}
\end{figure}

\begin{figure*}[t]
\centering
\includegraphics[width=0.93\textwidth]{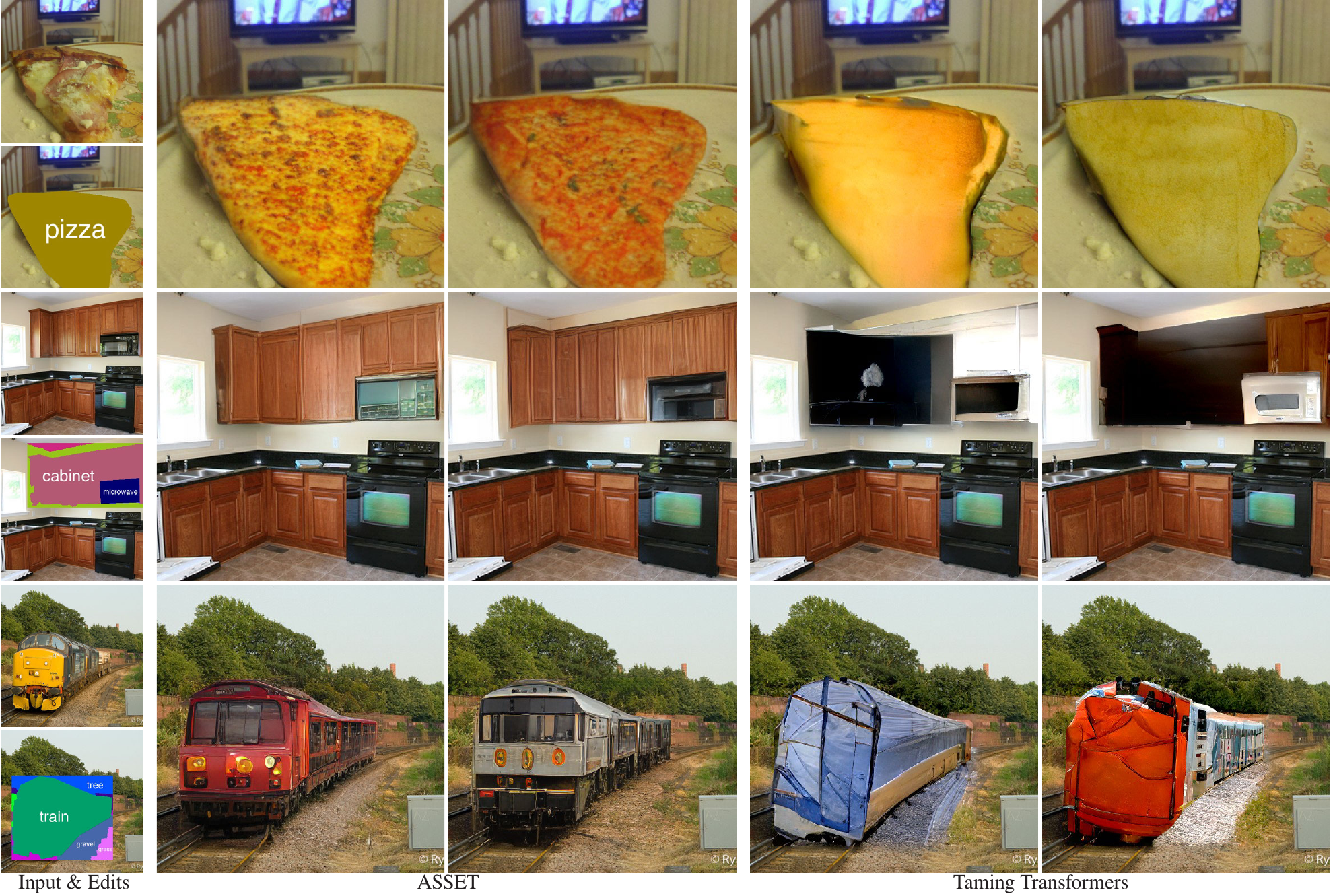}
\caption{
\rev{Qualitative results and comparisons with Taming Transformers \cite{esser2021taming} on COCO-Stuff at $512 \times 512$ resolution.}}
\label{fig:coco}
\end{figure*}

\begin{figure*}[t]
\centering
\includegraphics[width=0.93\textwidth]{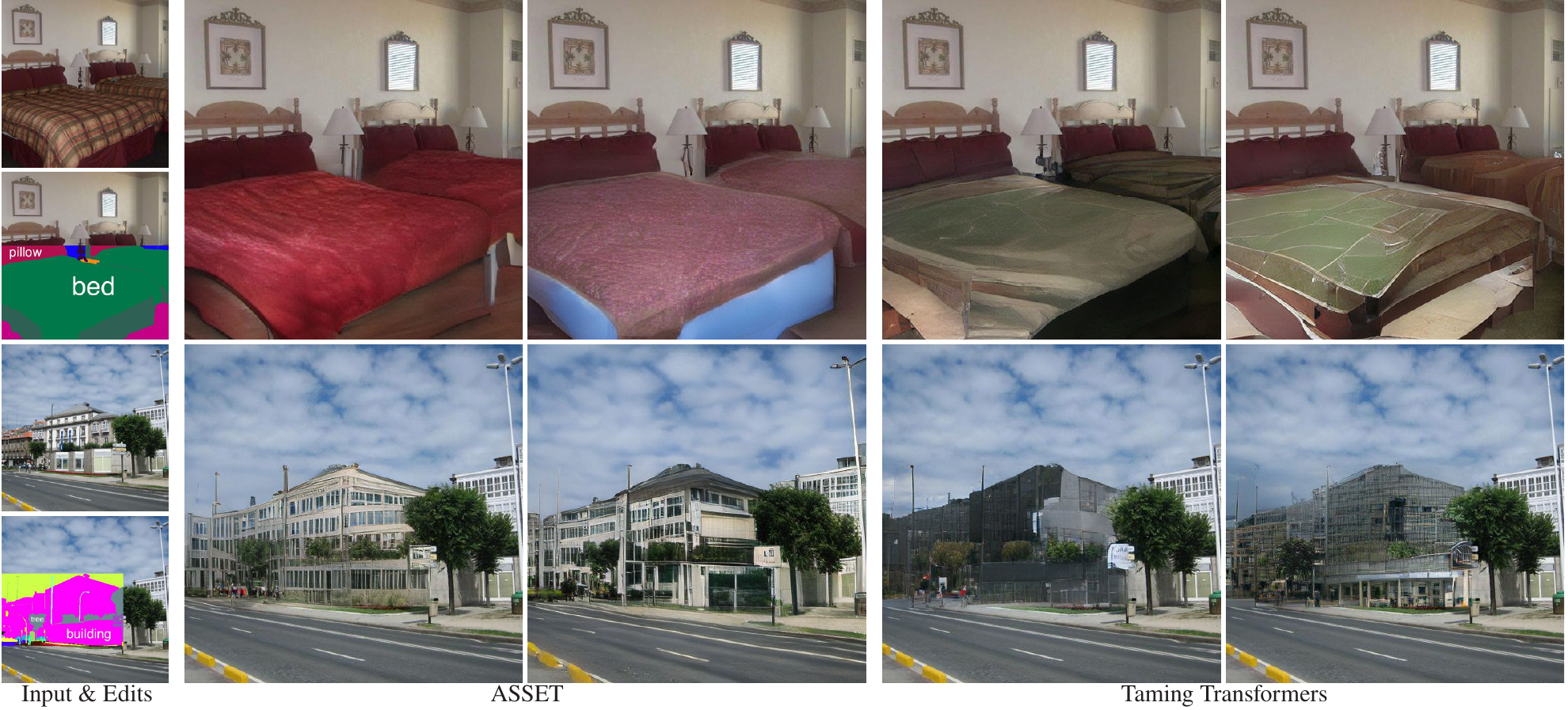}
\caption{
\rev{Qualitative results and comparisons with Taming Transformers \cite{esser2021taming} on ADE20K at $512 \times 512$ resolution.}}
\label{fig:ade}
\end{figure*}

\paragraph{Qualitative evaluation.}  
For qualitative evaluation,  a brush of a semantic class is used, painting over the image. Figure \ref{fig:comparison_256} shows comparison with all competing methods at $256 \times 256$ resolution.
Since we do not need SGA at small resolutions, we only use our guiding transformer for these examples.
Compared to other approaches, our method produces more coherent content with fewer artifacts.
In Figure \ref{fig:comparison_1024_TT} and Figure \ref{fig:comparison_1024_ImageBART}, \rev{we show the comparison on $1024 \times 1024$ images against Taming Transformers and ImageBART respectively.} Figure~\ref{fig:coco} and Figure~\ref{fig:ade} \rev{show qualitative results on COCO-Stuff and ADE20K at $512 \times 512$ resolution.} Even at high resolution, our method can synthesize coherent content across the whole image, while Taming Transformers and ImageBART fail to capture long-range dependency and sometimes ignore the user edits.

\begin{table}[t]
\renewcommand{\tabcolsep}{5pt}  
\renewcommand{\arraystretch}{1} 
\centering
\caption{Effects of different forms of attention on the Flickr-Landscapes dataset at $512\times512$ resolutions.}
\vspace{3mm}
\begin{tabular}{lccccccc}
\toprule
Method & LPIPS $\downarrow$ & FID $\downarrow$  & SSIM $\uparrow$ & mIoU $\uparrow$ & accu $\uparrow$ & div $\uparrow$ \\
\midrule
\emph{Sliding} & 0.214 & 10.7 & 0.847 & 52.6 & 63.1 & \textbf{0.180}  \\
\emph{Local} & 0.209 & 10.1 & 0.853 & 51.1 & 62.4 & 0.152 \\
\emph{Random} & 0.202  & 9.6 & 0.851 & 50.0 & 61.6 & 0.157 \\
\emph{ASSET} & \textbf{0.186} & \textbf{8.4} & \textbf{0.856} & \textbf{53.5} & \textbf{64.7} & 0.145 \\
\bottomrule
\end{tabular}

\label{table:ablation}
\end{table}

\paragraph{User study.} 
To further evaluate perceptual image quality we also conduct an Amazon MTurk study.
We showed participants a masked input image, along with a randomly ordered pair of images synthesized by ASSET and one of our baseline algorithms.
The participants were then asked which edited image looks more photo-realistic and coherent with the rest of the image.
Figure~\ref{fig:user_study} summarizes 1500 user responses for $256\times256$ and $1024\times1024$ resolutions.
The study shows that our method receives the most votes for better synthesis compared to other methods in both resolutions, with the largest margin at the highest $1024\times1024$ resolution.

\paragraph{Ablation study.}
To evaluate the effect of our SGA, we perform several ablations with different attention approaches.
The following variants are evaluated at $512\times512$ resolution:
\textbf{(1) \emph{Sliding}}: we use our guiding transformer with the sliding window approach as in \cite{esser2021taming}.
\textbf{(2) \emph{Local}}: we remove our top-$K$ attention and only use neighboring window attention $\N(r)$.
\textbf{(3) \emph{Random}}: we use random instead of top-$K$ attention similar to \cite{zaheer2020big}.
Table~\ref{table:ablation} shows the performance of all variants compared to our full model.
Our model outperforms all variants in all metrics except for diversity.
As before, we observe that higher diversity can be achieved at the cost of poorer image consistency.
In Figure~\ref{fig:ablation} we show qualitative comparisons with the proposed variants trained at $1024 \times 1024$ resolution.
As we can see, without the SGA component, the image consistency and perceptual quality decreases as the model either only attends to local areas (\emph{sliding} and \emph{local}) or fails to attend to important image regions at each sampling step (\emph{random}). 

\paragraph{\rev{Model capacity comparison with TT}} 
We compare the number of transformer parameters with TT in Table \ref{table:capacity}. Our transformer's number of parameters is ${\sim}12\%$ larger than the one in TT for the Landscape dataset, and much smaller for the COCO-Stuff and ADE20K datasets. 
We note that ASSET's and TT's CNNs have the same number of parameters.

\begin{figure}[t!]
  \centering
  \includegraphics[width=0.99\linewidth]{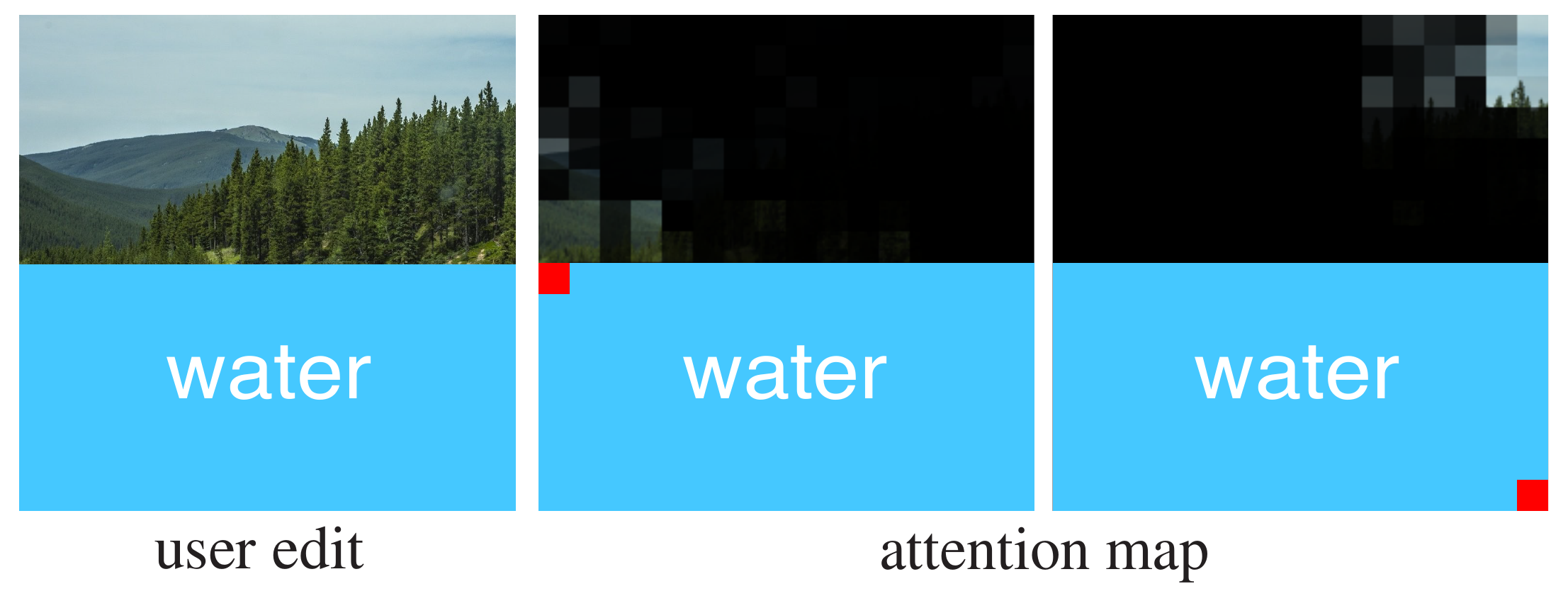}
  \caption{
  \rev{Encoder self-attention visualized for two different query points (shown as red). The image regions acquiring higher attention are the ones more relevant to generate water reflections at each of these points.} }
  \label{fig:vis}
\end{figure}

\paragraph{\rev{Attention visualization.}}
We use Attention Rollout \cite{abnar2020quantifying} to visualize the attention map of our guiding transformer encoder. Specifically, we average attention weights of the guiding transformer encoder across all heads and then recursively multiply the resulting averaged attention matrices of all layers. The attention maps for two different query points are presented in Figure \ref{fig:vis}. The guiding transformer can attend to informative regions for different query points. For the query points in Figure \ref{fig:vis}, the regions with high attention correspond to image areas useful to synthesize reflection of scenery at each of these points.

\begin{figure}[t!]
  \centering
  \includegraphics[width=0.99\linewidth]{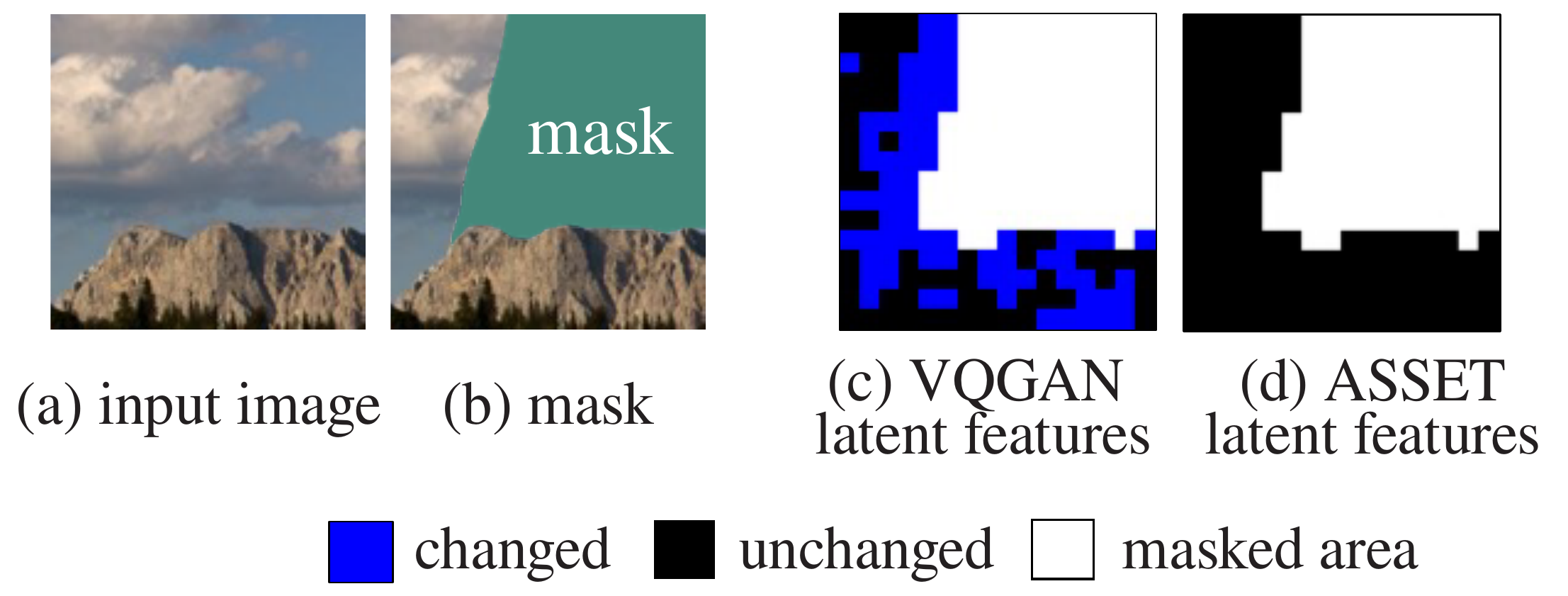}
  \caption{
  \rev{The masked area of the input image is replaced with random noise. The difference of $16 \times 16$ latent features between the original image (a) and the masked image (b) are visualized on the right. Changed and unchanged latent features are visualized in blue and black respectively. Image (c) shows how unmasked image latent tokens are affected (blue tokens) by the masked area in the original VQGAN. Image (d) shows that our image encoder successfully prevents leakage from the masked area to the unmasked area.}}
  \label{fig:leakage}
\end{figure}

\paragraph{\rev{VQGAN leakage visualization.}}
We employ partial convolutions \cite{liu2018image} and region normalization \cite{yu2020region} in our image encoder while processing the unmasked image regions.
The reason is that the features produced for unmasked regions should not be affected by the masked image regions. Partial convolutions and region normalization avoid any information leakage of the masked area. 
In Figure \ref{fig:leakage}, we visualize leakage for the original VQGAN and our improved image encoder. With our modification, the latent features produced for unmasked regions are independent of the masked area.

\begin{table}[tbp]
\renewcommand{\tabcolsep}{5pt}  
\centering
\caption{
\rev{Number of transformer parameters for TT and ASSET.} }
\begin{tabular}{lcc}
\toprule
Dataset  & TT & ASSET  \\
\midrule

Landscape  & \textbf{307M} & 343M  \\
COCO-Stuff  & 651M & \textbf{365M}  \\
ADE20K  & 405M & \textbf{269M}\\
\bottomrule
\end{tabular}

\label{table:capacity}
\end{table}

\paragraph{\rev{Inference speed comparisons.}}
Following \cite{cao2021image, esser2021imagebart}, we record the average inference time on Flickr Landscape and ADE20K as shown in Table \ref{table:inference_time}. The inference speed is influenced by the size of the masked region relevant to the size of the input image (i.e., ratio of the masked region). Following \cite{cao2021image}, we  report the average masked ratio in this table. Our method achieves similar inference speed with TT, while producing much higher-quality results than TT. 

\begin{table}[tbp]
\renewcommand{\tabcolsep}{5pt}  
\centering
\caption{
\rev{Average inference time in seconds per image.}}
\begin{tabular}{lcccc}
\toprule
Resolution & Dataset & Masked Ratio  & TT & ASSET  \\
\midrule

1024 & Landscape & 0.287 & \textbf{52.3} & 55.8  \\
512 & ADE20K  & 0.296 & 16.9 & \textbf{10.6}\\
\bottomrule
\end{tabular}

\label{table:inference_time}
\end{table}

\paragraph{\rev{Inference time of guiding transformer.}} Measured on the Landscape dataset, the average inference time of our guiding transformer ($256$ $\times$ $256$ resolution) represents only a small fraction ($3.4 \%$) of the total inference time of the full ASSET pipeline. The majority of the inference time ($96.6 \%$) is taken by our architecture operating at high resolution, which is the crucial part significantly accelerated by our SGA mechanism.

\paragraph{\rev{Comparison with full attention.}}
Based on an NVIDIA A100 ($40$GB VRAM) at $1024 \times 1024$ resolution with a batch size of $1$, the transformer architecture requirements with full attention exceeds the available memory during training. Using our Sparsified Guided Attention mechanism, the transformer architecture utilizes $37$GB at train time. In terms of inference time during testing, the cost of the guiding transformer is significantly lower: ASSET is about $20$ times faster at test time compared to using full attention at $1024$ $\times$ $1024$ resolution.

\section{Limitations and Conclusion}
We introduce a novel transformer-based approach for semantic image editing at high resolutions.
Previous approaches have difficulty in modeling long-range dependencies between image areas that are far apart, resulting in unrealistic and inconsistent image edits.
To this end, we introduce a novel attention mechanism called \emph{Sparsified Guided Attention} (SGA), which uses the full attention map at the coarse resolution to produce a sparse attention map at full resolution.
Our experiments show that SGA outperforms other variants of localized or sparse attention, and allows us to obtain realistic and diverse image edits even at high resolutions of $1024\times 1024$ pixels.

While our approach can perform consistent and diverse edits at high resolutions of up to $1024\times 1024$ pixels, there are still avenues for further improvements.
A common issue in transformers including ours is that directly applying a trained model to generate content at a higher resolution degrades performance, since the learned positional embedding cannot adapt to the new resolution.
In addition, the repeated autoregressive sampling takes several minutes to perform a full edit at $1024 \times 1024$ resolution.
To alleviate this issue, we can sample a diverse set of outputs for a given edit in parallel on multiple GPUs.
Finally, the synthesized content may not be perfectly aligned with the provided mask since the masking takes place at a low resolution in the latent space.

\begin{acks}
This work is funded by Adobe Research.
\end{acks}

\bibliographystyle{ACM-Reference-Format}
\bibliography{references}
\appendix

\section{Implementation Details}
Here we provide implementation details of our network architecture and training procedure.
Our model is implemented in PyTorch.

\paragraph{Image encoder / decoder.}
Our image encoder (Section~\ref{sec:image_encoder}) and decoder (Section \ref{sec:image_decoder}) 
use the  architecture  shown  in  Table \ref{tab:vqarchitecture}. The design of the networks follows the architecture proposed in VQGAN \cite{esser2021taming}. One difference is that we employ partial convolutions \cite{liu2018image} and region normalization \cite{yu2020region} in our image encoder while processing the unmasked image regions.
The reason is that the features produced for unmasked regions should not be affected by the masked image regions. Partial convolutions and region normalization avoid any information leakage of the masked area.

The semantic map encoder follows the same architecture with regular convolutions.
We note that during training, VQGAN measures the reconstruction error in terms of both the image and semantic map. Thus, along with the decoder for the image, we also use a decoder to reconstruct the semantic map. The semantic map decoder uses an architecture following the ``decoder'' column in Table \ref{tab:vqarchitecture}. A minor difference is that  the number of input/output channels is changed from $3$ to the number of categories $C$ in the semantic map encoder and decoder.

\begin{table}[tbp]
\renewcommand{\tabcolsep}{4.5pt}  
\centering
\caption{Ablation for use of global blocks at $512\times512$ resolution.}
\begin{tabular}{lcccccc}
\toprule
Method & LPIPS $\downarrow$ & FID $\downarrow$  & SSIM $\uparrow$ & mIoU $\uparrow$ & accu $\uparrow$ \\
\midrule

\emph{R+G} & 0.207 & 9.5 & 0.849 & 50.3 & 61.7  \\
\emph{Random} & 0.202  & 9.6 & 0.851 & 50.0 & 61.6  \\
\emph{ASSET} & \textbf{0.186} & \textbf{8.4} & \textbf{0.856} & \textbf{53.5} & \textbf{64.7}  \\
\bottomrule
\end{tabular}
\label{table:supp_ablation}
\end{table}

\paragraph{Transformer.}
Our transformer (Section~\ref{sec:transformer}) follows the architecture presented in BART \cite{lewis2019bart}.
All the hyperparameters for the transformer are described in Table \ref{tab:bartarchitecture}.

Following \cite{chu2021conditional, chu2021twins}, the position encoding generator (PEG) is placed before each transformer encoder layer. The position encoding generator is a $5 \times 5$ depth-wise convolution with the padding size of $2$, which convolves with each block independently. 
Similar to the encoder layer, we add one PEG before the first decoder layer which takes the encoder output representation as input to produce positional embeddings for the transformer decoder.

\paragraph{Sparsified Guided Attention.}
For each transformer layer and attention head, a full attention matrix 
\mbox{$\bA_{low} \in \mathds{R}^{256\times 256}$} is computed from the downsampled input image. The computation of the block attention matrix $\bB$ for high-resolution is guided by $\bA_{low}$. Specifically, in our experiments, the number of blocks $N$ is set to $64$. Each block consists of $\frac{256}{64} = 4$ tokens at low-resolution. The affinity value of each block corresponds to a $4\times4$ region in $\bA_{low}$. In our implementation, we use a 2D average pooling layer with kernel size $4$ and stride $4$ to downsample $\bA_{low}$ into $\bB$.

\paragraph{Training details.}
For the image encoder/decoder and semantic encoder/decoder, we used the Adam optimizer \cite{kingma2014adam} with learning rate $7.2 \cdot 10^{-6}$ and batch
size $16$. During the training of the guiding transformer, we used the AdamW optimizer \cite{loshchilov2017decoupled} with learning rate $3.2 \cdot 10^{-5}$ and batch
size $224$. During the finetuning of the SGA-transformer, we used the AdamW optimizer \cite{loshchilov2017decoupled} with learning rate $1.2 \cdot 10^{-5}$ and batch size $8$. All training is done on $8$ A100 GPUs.

\vqarchitecture
\bartarchitecture

\section{Additional Results and Comparisons}
 
\begin{figure*}[t]
\centering
\includegraphics[width=0.93\textwidth]{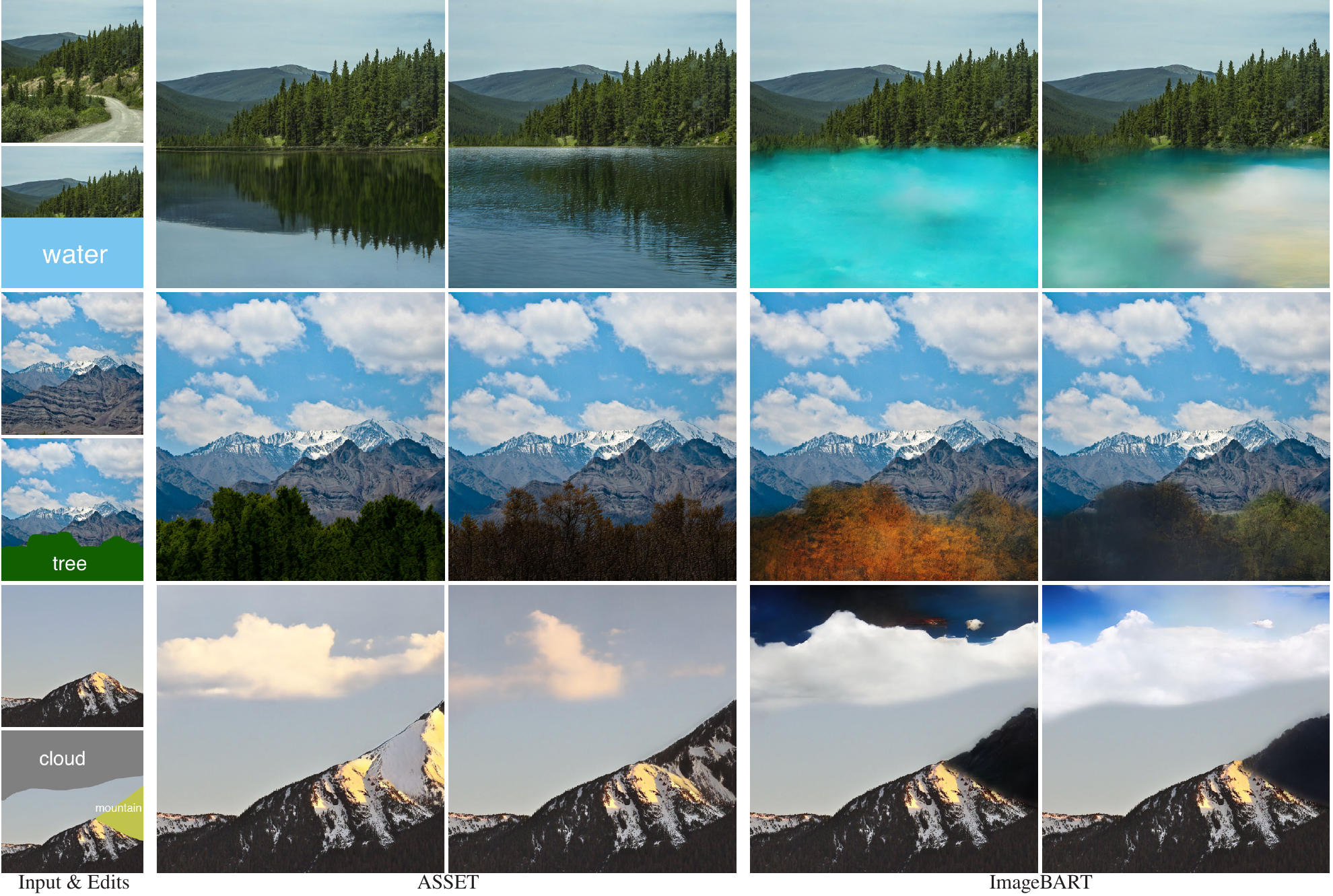}
\caption{Comparison of ASSET with ImageBART \cite{esser2021imagebart} at $1024 \times 1024$ resolution.}
\label{fig:supp_1024_ImageBART}
\end{figure*}

\begin{figure*}[t]
\centering
\includegraphics[width=0.93\textwidth]{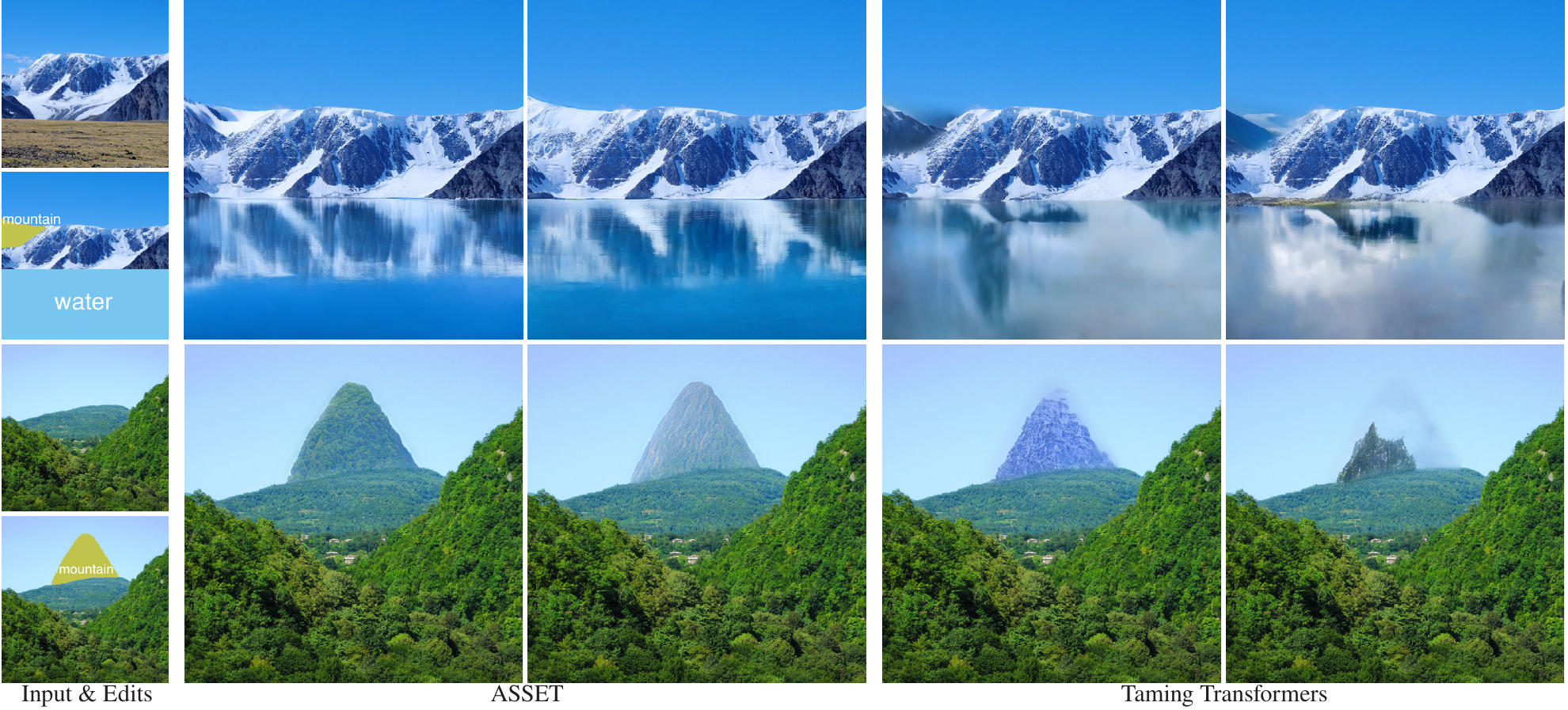}
\caption{
\rev{Comparison of ASSET with Taming Transformers \cite{esser2021taming} at $1024 \times 1024$ resolution.} }
\label{fig:supp_1024_TT}
\end{figure*}

\paragraph{Additional comparisons}
Please see Figure \ref{fig:supp_1024_ImageBART} and Figure \ref{fig:supp_1024_TT} for more comparisons at $1024$ resolution.

\begin{figure}[t!]
  \centering
  \includegraphics[width=0.99\linewidth]{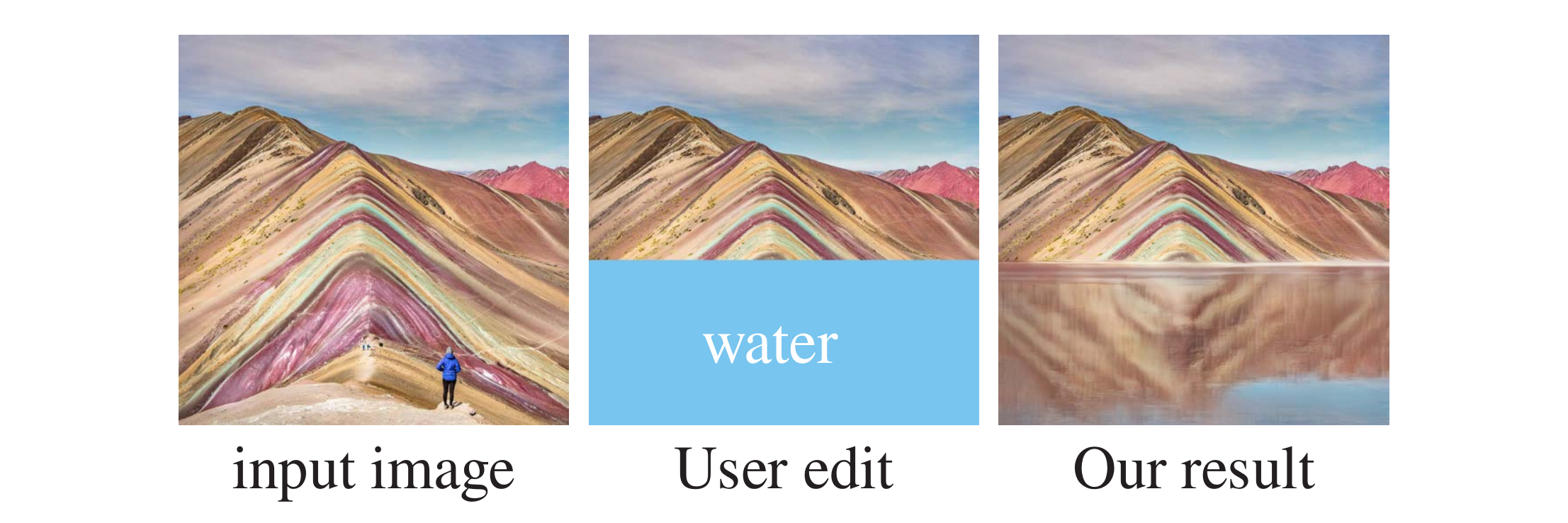}
  \vspace{-3mm}
  \caption{\rev{Example of a less successful result.}}
  \vspace{-3mm}
  \label{fig:failure}
\end{figure}

\paragraph{Adding global blocks.}
In our ablation study, we also experimented with the global block presented in BigBird \cite{zaheer2020big}. Specifically, we make the first and last blocks ``global'', which attend over the entire sequence. Similar to \cite{zaheer2020big}, we use the global attention together with the local attention and random attention -- this variant is referred to as \emph{R+G}. The results did not improve compared to the \emph{Random} variant in terms of our evaluation metrics (see Table \ref{table:supp_ablation}).

\rev{
\paragraph{Failure case.}
Structured textures such as the mountain in Figure \ref{fig:failure} is challenging for reflection synthesis. In this case,
our result may not reproduce the texture well.}

\clearpage
\end{document}